\documentclass[10pt,twocolumn,letterpaper]{article}

\usepackage[pagenumbers]{cvpr} 

\newcommand{\paren}[1]{\left( #1 \right)}
\newcommand{\parens}[1]{( #1 )}
\newcommand{\bracket}[1]{\left[ #1 \right]}
\newcommand{\normtwo}[1]{\left\lVert #1 \right\rVert}

\newcommand{\Paragraph}[1]{\noindent \textbf{#1}}
\usepackage{appendix}
\usepackage{titling}
    \pretitle{\vspace*{-3pt}\begin{center}\Large \bf}
    \posttitle{\vspace*{10pt}\par\end{center}}
    \preauthor{\large\begin{center}\par}
    \postauthor{\vspace*{6pt}\par\end{center}}
    \predate{}
    \date{}
    \postdate{}
    
\usepackage{graphicx}
\usepackage{amsmath}
\usepackage{amssymb}
\usepackage{multirow}
\usepackage{cancel}
\usepackage{enumerate}
\usepackage{graphics}
\usepackage{comment}
\usepackage{stackengine}
\usepackage{pifont}
\newcommand*\samethanks[1][\value{footnote}]{\footnotemark[#1]}

\usepackage{booktabs}
\usepackage{tabularx}
\usepackage[pagebackref,breaklinks,colorlinks]{hyperref}

\usepackage[capitalize]{cleveref}
\crefname{section}{Sec.}{Secs.}
\Crefname{section}{Section}{Sections}
\Crefname{table}{Table}{Tables}
\crefname{table}{Tab.}{Tabs.}

\def\cvprPaperID{5516} 
\def\confName{CVPR}
\def\confYear{2022}

\begin{document}

\title{CVF-SID: Cyclic multi-Variate Function \textit{for}\\ Self-Supervised Image Denoising by Disentangling  Noise from Image}

\author{Reyhaneh Neshatavar$^{1}$\thanks{equal contribution} \qquad Mohsen Yavartanoo$^{1}$\samethanks \qquad Sanghyun Son$^{1}$ \qquad Kyoung Mu Lee$^{1,2}$ \\$^{1}$Dept. of ECE \& ASRI, $^{2}$IPAI, Seoul National University, Seoul, Korea\\
{\tt\small \{reyhanehneshat,myavartanoo,thstkdgus35,kyoungmu\}@snu.ac.kr}}
\maketitle

\begin{abstract}
Recently, significant progress has been made on image denoising with strong supervision from large-scale datasets.
However, obtaining well-aligned noisy-clean training image pairs for each specific scenario is complicated and costly in practice.
Consequently, applying a conventional supervised denoising network on in-the-wild noisy inputs is not straightforward.
Although several studies have challenged this problem without strong supervision, they rely on less practical assumptions and cannot be applied to practical situations directly.
To address the aforementioned challenges, we propose a novel and powerful self-supervised denoising method called CVF-SID based on a \textbf{C}yclic multi-\textbf{V}ariate \textbf{F}unction (CVF) module and a self-supervised image disentangling (SID) framework.
The CVF module can output multiple decomposed variables of the input and take a combination of the outputs back as an input in a cyclic manner.
Our CVF-SID can disentangle a clean image and noise maps from the input by leveraging various self-supervised loss terms.
Unlike several methods that only consider the signal-independent noise models, we also deal with signal-dependent noise components for real-world applications.
Furthermore, we do not rely on any prior assumptions about the underlying noise distribution, making CVF-SID more generalizable toward realistic noise.
Extensive experiments on real-world datasets show that CVF-SID achieves state-of-the-art self-supervised image denoising performance and is comparable to other existing approaches.
The code is publicly available from this \href{https://github.com/Reyhanehne/CVF-SID_PyTorch}{link}.
\end{abstract}

\section{Introduction}

Image denoising is an active research topic and has attracted increasing attention due to its practicality in computer vision.
The fundamental idea of image denoising is to remove unwanted noise signals from a given input and restore a noise-free clean image.
Following the recent advances in convolutional neural networks (CNN), the latest denoising methods have achieved dramatic performance compared to the traditional algorithms.
Specifically, those methods resort to supervised learning on the large-scale synthetic dataset, where noise is simply modeled with additive white Gaussian (AWGN)~\cite{sparse, 6126278, 6909762, DBLP:journals/corr/Lefkimmiatis16, DBLP:journals/corr/ZhangZCM016}.

Nevertheless, recent studies~\cite{DBLP:journals/corr/abs-1807-04686, DBLP:journals/corr/abs-1904-07396} have observed that the denoising models learned on synthetic images do not generalize well on practical examples.
The primary reason for this issue is that real-world noise distribution differs from the synthetic AWGN.
To deal with this limitation, few attempts have been made to acquire realistic noisy-clean image pairs~\cite{8578280} in the wild.
Still, this process is challenging and sometimes unavailable as it requires multiple shots under the same static scene with several constraints.

\begin{figure}[t!]
    \centering
    \captionsetup[subfigure]{labelformat=empty}
    \renewcommand{\wp}{0.33\linewidth}
    \subfloat[\scriptsize Input Noisy]{\includegraphics[width=\wp,page=1]{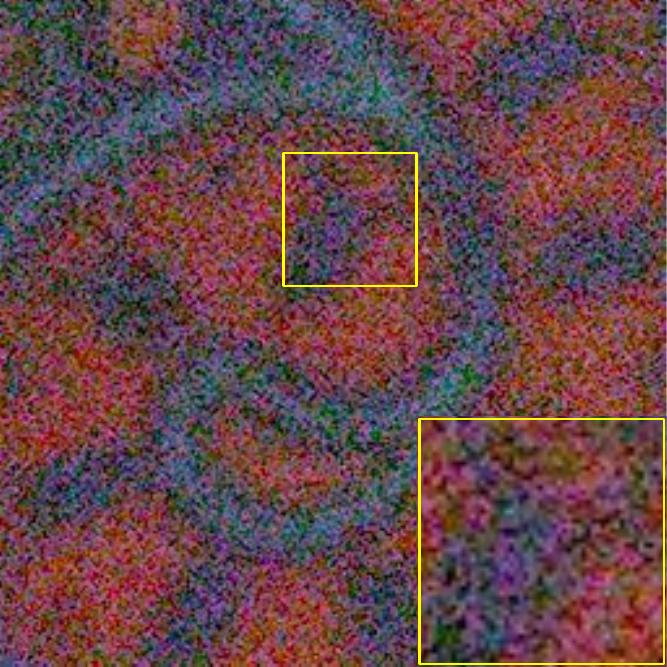}}
    \hfill
    \subfloat[\scriptsize Ground Truth]{\includegraphics[width=\wp,page=2]{figures/image_607.pdf}}
    \hfill
    \subfloat[\scriptsize N2V~\cite{DBLP:journals/corr/abs-1811-10980}: 24.00dB]{\includegraphics[width=\wp,page=3]{figures/image_607.pdf}}
    \\
    \subfloat[\scriptsize N2S~\cite{DBLP:journals/corr/abs-1901-11365}: 25.34dB]{\includegraphics[width=\wp,page=4]{figures/image_607.pdf}}
    \hfill
    \subfloat[\scriptsize R2R~\cite{Pang_2021_CVPR}: 30.37dB]{\includegraphics[width=\wp,page=5]{figures/image_607.pdf}}
    \hfill
    \subfloat[\scriptsize \textbf{CVF-SID (Ours): 32.99dB} ]{\includegraphics[width=\wp,page=7]{figures/image_607.pdf}}
    \\
    \vspace{-2mm}
    \caption{
        \textbf{Real-world image denoising results on the SIDD validation dataset.}
        %
        %
        In contrast to R2R, our CVF-SID is directly applicable to sRGB images. 
        N2V and N2S fail to restore the clean image, and R2R loses the details evidenced in the yellow patch.
        %
    }
    \label{fig1}
    \vspace{-4mm}
\end{figure}

Advanced methods overcome the lack of paired images by adopting novel un-/self-supervised frameworks.
The generation-based approaches~\cite{Hong_Fan_Jiang_Feng_2020, Jang_2021_ICCV} utilize unsupervised adversarial training.
They first generate noisy samples from a set of clean images by imitating the noise distribution of the target dataset.
Then, a denoising model can be trained in a supervised manner with the generated noisy-clean image pairs.
Nevertheless, they require appropriate clean images that are not always available due to domain differences between noisy and clean images.
Rather than using clean ground-truth data, Noise2Noise (N2N)~\cite{DBLP:journals/corr/abs-1803-04189} uses two noisy images taken from the same scene and configuration.
While N2N shows comparable performance with the supervised methods, it is less practical as multiple noisy images under the same scene are required.

As an alternative solution, several strategies~\cite{DBLP:journals/corr/abs-1811-10980, DBLP:journals/corr/abs-1901-11365, moran2019noisier2noise, Quan_2020_CVPR, huang2021neighbor2neighbor, Pang_2021_CVPR} have been proposed to train their methods on noisy images only.
To generate feasible input-target pairs from a single noisy image, these recent approaches try to synthesize two independent noisy images from the input.
However, these methods are usually constructed by assuming a specific distribution, \eg, AWGN, or less practical configurations for the underlying noise.
Such an assumption limits their practical applications where the assumption does not hold.
For instance, recent Recorrupted-to-Recorrupted (R2R)~\cite{Pang_2021_CVPR} is not applicable to sRGB inputs directly, while digital images are usually stored using sRGB color space.
Also, this method requires additional prior knowledge, \eg, a pre-trained model with provided noise level function~(NLF)~\cite{1640848} by Raw-RGB images, on real-world noise, which is not trainable using sRGB images only.

To mitigate the limitations mentioned above, we present a self-supervised denoising method for the real-world sRGB images.
To this end, we propose a novel cyclic multi-variate function (CVF), which disentangles its input into several sub-components and retakes a combination of its outputs as an input.
We utilize CVF to design our self-supervised image disentangling model (CVF-SID) for sRGB image denoising.
Under various self-supervised training objectives, our CVF-SID can learn to disentangle the noise-free image, signal-dependent and signal-independent noises from a given noisy sRGB input.
Furthermore, we propose a self-supervised data augmentation strategy for CVF-SID to effectively increase the number of training samples.
%
Our main contributions can be summarized as follows:
\begin{itemize}
    \item We introduce CVF-SID, a novel self-supervised method for image denoising based on our defined cyclic multi-variate function (CVF).
    CVF-SID disentangles a given real-world noisy input to clean image, signal-dependent, and signal-independent noises.
    \item For fully self-supervised CVF-SID, we propose various training objectives and an augmentation strategy.
    \item Experimental results demonstrate that our CVF-SID achieves superior denoising performance among several un-/self-supervised methods on real-world sRGB images as shown in Figure~\ref{fig1} and is comparable with the other approaches.
\end{itemize}
\begin{figure}
\centering
    \includegraphics[width=\linewidth]{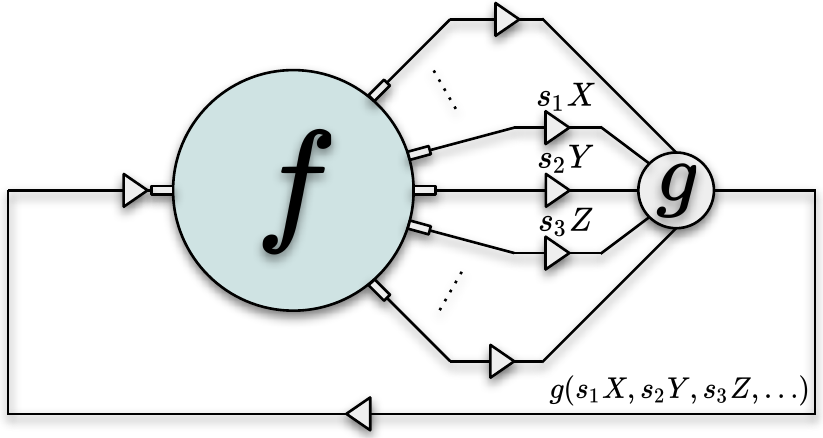}
    \vspace{-2mm}
    \caption{
        \textbf{The proposed Cyclic multi-Variate Function (CVF)}. Our CVF $f$ takes a combination $g \paren{s_1 X, s_2 Y, s_3 Z,\dots}$ of multiple variables as an input and outputs the decomposed variables.}
    \label{fig:cvf}
    \vspace{-4mm}
\end{figure}

\section{Related work}
%
\begin{figure*}
\centering
    \subfloat[]{\includegraphics[width=0.49 \linewidth, page=1]{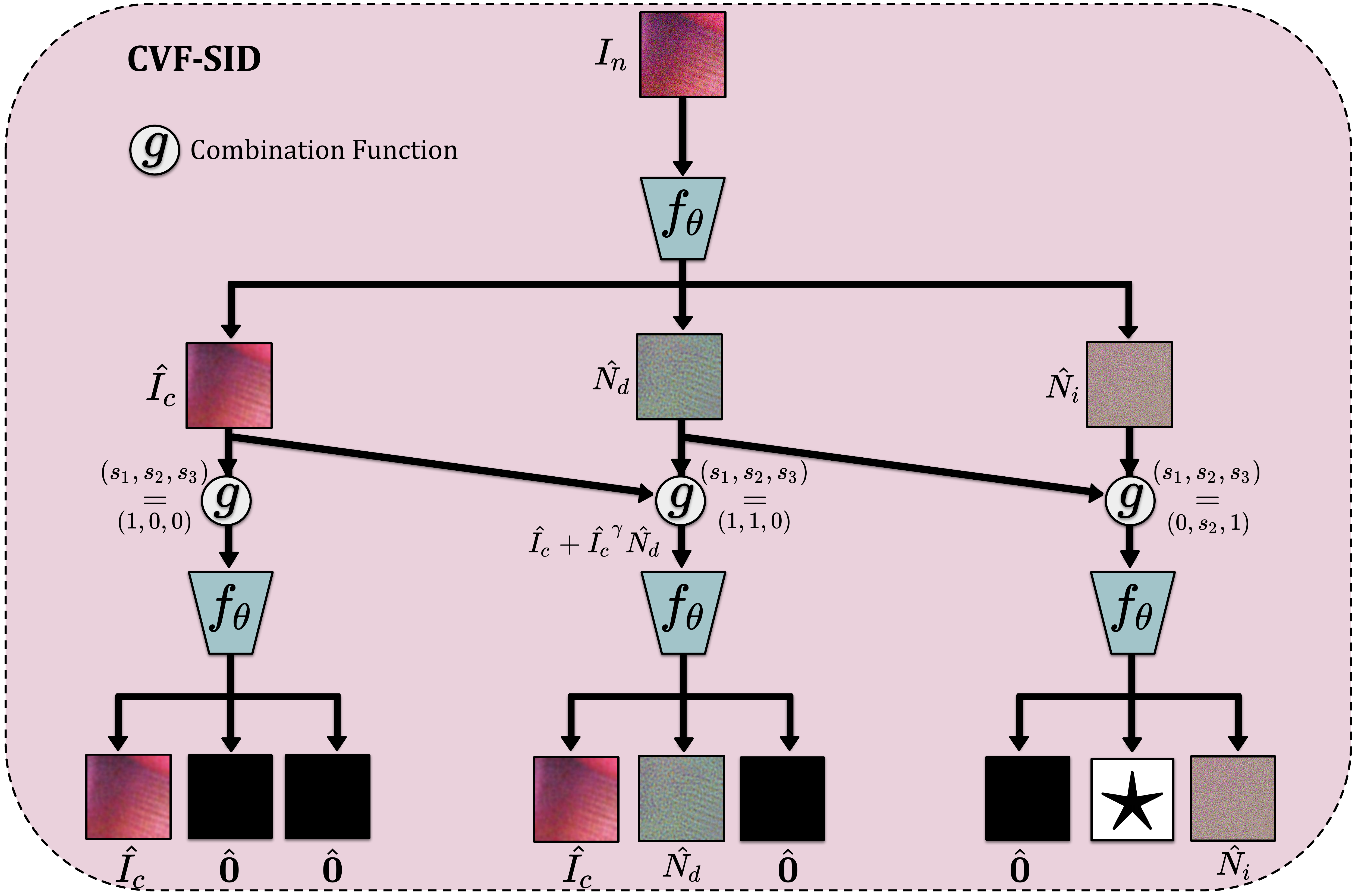}\label{fig:framework_a}}
    \hfill 
    \subfloat[]{\includegraphics[width=0.49\linewidth, page=2]{figures/network/framework.pdf}\label{fig:framework_b}}
    \vspace{-2mm}
    \caption{
        \textbf{Overview of our CVF-SID framework.}
        (a) We employ the network $f_{\theta}$ to disentangle the clean image, signal-dependent, and signal-independent noises from a noisy input image.
        In our cyclic network, we feed the outputs of our network again to the network to ensure that each output is pure and does not have the information of the other outputs.
        (b) We also feed some self-supervised augmented images to our network to better estimate the distribution of each noise and improve the performance.
    }
    \label{fig:framework}
    \vspace{-4mm}
\end{figure*}

Traditional approaches on image denoising such as NLM~\cite{1467423}, BM3D~\cite{sparse}, WNNM~\cite{6909762}, NC~\cite{ipol.2015.125}, K-SVD~\cite{DBLP:journals/corr/abs-1909-13164}, or EPLL~\cite{Hurault_2018} adopt non-learning-based formulation.
Recently, deep learning has demonstrated impressive performance on image denoising.
In general, they can be categorized based on the way how the network is trained.
\subsection{Training on paired noisy-clean images}
In general, supervised denoising networks are trained on synthetic noisy-clean images, where the noise is assumed to be additive white Gaussian (AWGN) of a certain level~\cite{burger2012image, 7527621}.
DnCNN~\cite{DBLP:journals/corr/ZhangZCM016} is the first CNN-based approach for image denoising, which introduces residual learning and outperforms the traditional methods.
FFDNet~\cite{ffdnet} further proposes a fast and flexible solution to handle various noise levels within a single model by taking a noise map as an additional input.
However, the conventional methods do not generalize well on real-world applications due to domain discrepancy between realistic and synthetic noise.
To overcome this limitation, several approaches such as CBDNet~\cite{DBLP:journals/corr/abs-1807-04686}, RIDNet~\cite{DBLP:journals/corr/abs-1904-07396}, or DIDN~\cite{9025411} train their methods on realistic noisy-clean pairs~\cite{8578280}.
Nevertheless, gathering well-aligned noisy-clean pairs from the real-world scenes is challenging and not very practical, as it requires huge human labor under controlled environments~\cite{DBLP:journals/corr/PlotzR17}.

\subsection{Training on unpaired noisy-clean images}
To overcome the limitations of the supervised approaches, generation-based methods aim to synthesize noisy samples from clean images in an unsupervised manner~\cite{goodfellow2014generative}. 
They first try to simulate realistic noise in the adversarial training framework and then train a denoising model on the generated noisy-clean pairs.
GCBD~\cite{Chen2018ImageBD} is the first generation-based method for blind denoising. 
However, it is not applicable to real-world scenarios since the method considers additive noise only, while real-world noise is not.
Further, UIDNet~\cite{Hong_Fan_Jiang_Feng_2020} employs an image sharpening technique to estimate arbitrary noise distribution in real-world cases.
Recent C2N~\cite{Jang_2021_ICCV} tries to explicitly consider signal-independent, dependent, and spatially-correlated noise in their generation framework.
However, such methods require clean images to generate the corresponding noisy images, which is not applicable when the scene distribution of the noisy images is not matched to the existing clean samples.

\subsection{Training on paired noisy-noisy images}
To alleviate the issues of generation-based approaches, some researchers tried to train their network on paired noisy-noisy images instead of paired noisy-clean or unpaired noisy-noisy images in an un-/self-supervised manner.
Noise2Noise (N2N)~\cite{DBLP:journals/corr/abs-1803-04189}, as a weakly supervised learning on image denoising, proposes to use several noisy images instead of ground truth images.
While it can achieve comparable results with supervised methods, taking several independent noisy images from the same scene is also very difficult in real-world cases.
Noise2Void (N2V)~\cite{DBLP:journals/corr/abs-1811-10980} and Noise2Self (N2S)~\cite{DBLP:journals/corr/abs-1901-11365} use a blind-spot learning to avoid learning the identity function without availability of paired images. 
These approaches remove the center pixel of each receptive field and predict it with other neighboring pixels. 
However, ignoring some pixels leads to the loss of some useful information and reduces the performance~\cite{8578280}.

Noisier2Noise~\cite{moran2019noisier2noise} is designed to handle spatially correlated noises by learning on noisy-noisy pairs.
Nevertheless, it requires additional information, \eg, the underlying noise distribution, which is a less practical assumption to be applied to real-world inputs.
Self2Self (S2S)~\cite{Quan_2020_CVPR} is proposed on blind denoising to generate paired data from a single noisy image by applying Bernoulli dropout. 
Later, Neighbor2Neighbor~\cite{huang2021neighbor2neighbor} proposes to create sub-sampled paired images based on pixel-wise independent noise assumption.
Recorrupted-to-Recorrupted (R2R)~\cite{Pang_2021_CVPR} expands the concept of Noisier2Noise~\cite{moran2019noisier2noise} toward real-world scenarios. 
However, R2R resorts to the Gaussian noise assumption when no raw information is provided for a given noisy input image, which prevents its practical applications on digital noisy sRGB images.
In contrast, our CVF-SID method can be trained on noisy sRGB images directly without generating pseudo noisy-noisy pairs.

\section{Method}
We introduce the concept of our CVF and construct our self-supervised denoising model, CVF-SID, on the formulation.
For convenience, we denote clean and noisy images as $I_c, I_n \in \mathbb{R}^{H \times W}$, respectively, where the image has a spatial resolution of $H \times W$.
Color channels, i.e., RGB, are omitted for simplicity.
We represent signal-dependent and independent noise maps as $N_d$ and $N_i$, respectively, where they have the same dimension to $I_n$.
\subsection{Cyclic multi-Variate Function}
\label{ssec:cvf}
We define a cyclic multi-variate function (CVF) $f$ as a mapping from $g \paren{X, Y, Z, \dots}$ to $\bracket{X, Y, Z, \dots}$, where $X$, $Y$, $Z$, $\dots$ are vectors and $g$ is a combination function.
Therefore, the function can take its outputs as an input again by combining the output values, as shown in Figure~\ref{fig:cvf}.
Also, for a set of scalar values $\bracket{ s_1, s_2, s_3, \dots }$, a decomposition of $g \paren{ s_1 X, s_2 Y, s_3 Z, \dots }$ should be $\bracket{ s_1 X, s_2 Y, s_3 Z, \dots }$.
By utilizing the aforementioned attributes of CVF, we aim to learn a denoising model in a self-supervised manner.
\subsection{Self-supervised image denoising using CVF}
In general, a noisy image $I_{n}$ can be expressed as a function of the clean image $I_c$, signal-dependent noise map $N_d$, and the signal-independent noise map $N_i$~\cite{torricelli2002modelling} as follows:
\begin{equation}
\label{eq:g_function}
I_{n} = I_{c} + I_{c}^{\gamma} N_{d} + N_{i},
\end{equation}
where $\gamma$ is a parameter regarding the correlation between the signal and the corresponding noise term.
Previous methods~\cite{DBLP:journals/corr/ZhangZCM016, Quan_2020_CVPR} have designed their model to take a single noisy image and reconstruct its clean counterpart, \ie, $f \paren{ I_n } = I_c$.
In contrast, we utilize the concept of CVF to design our network $f_\theta$ with learnable parameters $\theta$ for disentangling the given noisy image $I_n$ into the aforementioned three components $I_c$, $N_d$, and $N_i$ as shown in Figure~\ref{fig:framework_a}.
Following the notation in Section~\ref{ssec:cvf}, we denote the noisy image $I_n$ as $g \paren{ s_1I_c, s_2N_d, s_3N_i }$, where $s_1 = s_2 = s_3 = 1$.

\begin{figure}
\centering
    \includegraphics[width=\linewidth]{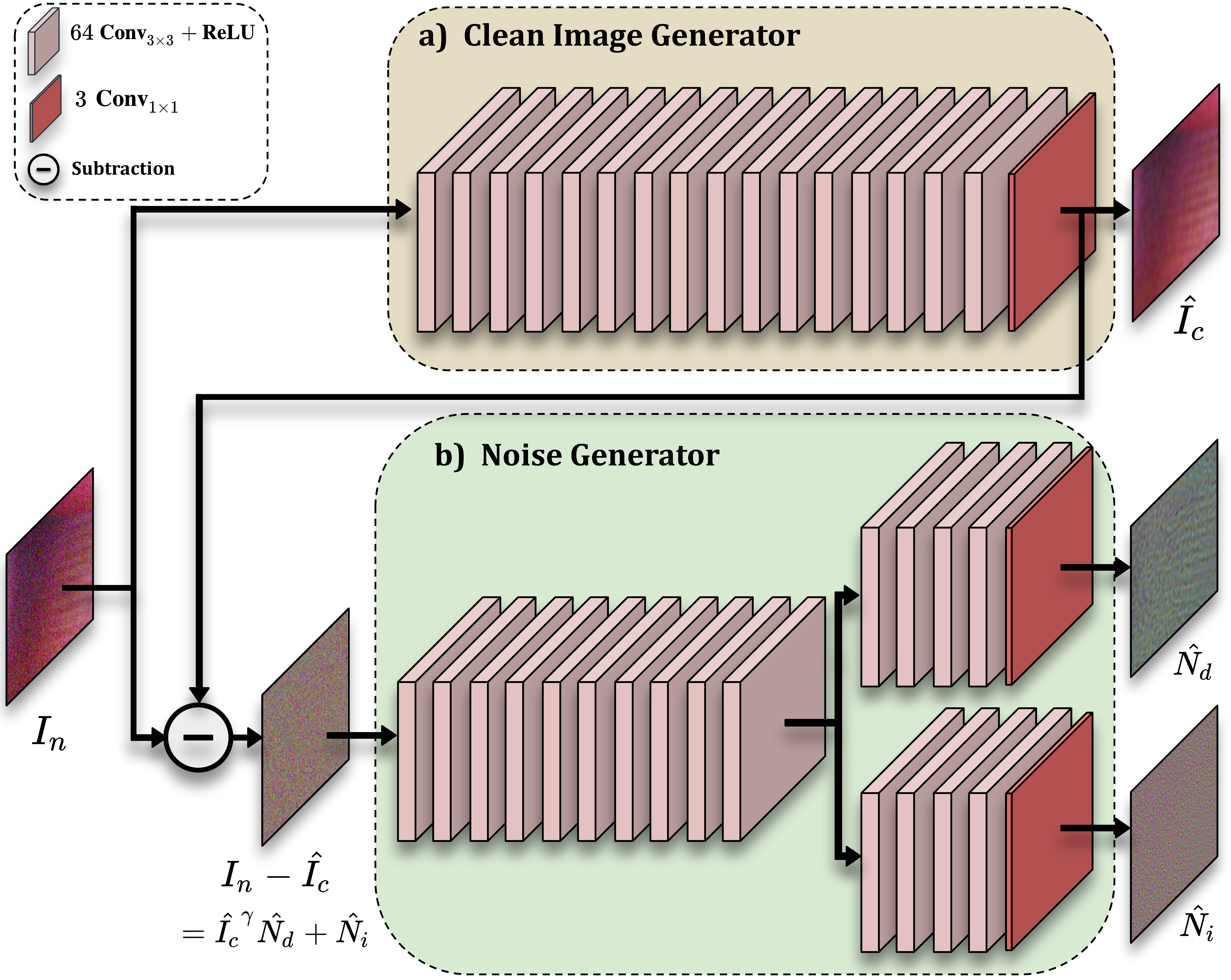}
    \vspace{-2mm}
    \caption{
        \textbf{Overview of our network architecture $f_{\theta}$.}
    }
    \label{fig:network}
    \vspace{-4mm}
\end{figure}

After decomposing the given noisy image into the clean image and the noise components, we exploit their statistical properties to construct a self-supervised cycle.
For practical reasons, we assume that the noise maps $N_d$ and $N_i$ are zero-mean~\cite{Liu2014PracticalSN} and spatially invariant with unknown distributions.
Moreover, we suppose that two elements in each pairs of $\paren{ I_{c}, N_{d} }$, $\paren{ I_{c}, N_{i} }$, and $\paren{ N_{d}, N_{i} }$ are independent.
In other words, $I_{c}$, $N_{d}$, and $N_{i}$ do not contain the information of each other.
We note that $N_d$ is a spatially invariant noise map that acts as a scaling factor in the signal-dependent noise term $I_c^{\gamma}N_d$. 

Under such assumptions, $f_\theta$ decomposes a given noisy image $I_n$ as follows:
\begin{equation}\label{eq:f_function}
\begin{split}
    f_{\theta} \paren{ I_{n} } &= \bracket{
        f_\theta^\text{clean} \paren{I_n},
        f_\theta^\text{dep} \paren{I_n},
        f_\theta^\text{indep} \paren{I_n}
    } \\
    &= \bracket{ \hat{I}_{c}, \hat{N}_{d}, \hat{N}_{i} },
\end{split}
\end{equation}
where $\hat{I}_{c}$, $\hat{N}_{d}$, and $\hat{N}_{i}$ denotes predicted clean image, signal-dependent noise, and signal-independent noise terms, respectively.
Since we do not use any noisy-clean image pairs, it is not possible to apply direct supervisions to the outputs of the function, and it is not guaranteed that $f_\theta$ disentangle these components perfectly.

Therefore, to train our model in a self-supervised manner, we feed the outputs $\hat{I}_{c}$, $\hat{N}_{d}$, and $\hat{N}_{i}$ again to the same network $f_\theta$ with shared parameters $\theta$.
Then, we constrain the second outputs, \eg, $f_\theta \parens{ \hat{I}_c } $, based on their desired properties.
Since a predicted noise-free image $\hat{I}_c$ should not contain any noise, regardless of the signal dependency, it can be modeled as $\hat{I}_c = g(\hat{I}_c, \mathbf{0}, \mathbf{0})$, where $s_1=1$ and $s_2=s_3=0$.
Here, we use $\mathbf{0}$ to represent an $H \times W$ array of zeros.
Therefore, given a predicted clean image $\hat{I}_c$ as an input, our $f_\theta$ has to generate the following outputs:
%
\begin{equation}\label{eq:input_clean}
    f_{\theta} \parens{ \hat{I}_{c} } = \bracket{ \hat{I}_{c}, \hat{\mathbf{0}}, \hat{\mathbf{0}} },
\end{equation}
%
where the predicted output noise maps $\hat{N}_d$ and $\hat{N}_i$ should be zeros which are denoted as $\hat{\mathbf{0}}$.

Our another assumption is derived from the observation that an image corrupted by the signal-dependent noise only should be decomposed to $\hat{I}_c$ and $\hat{N}_d$ or equivalently $\hat{I}_{c} + \hat{I}_{c}^{\gamma}\hat{N}_{d} = g(\hat{I}_c,\hat{N}_d,\mathbf{0})$, where $s_1=s_2=1$ and $s_3=0$.
Then, the network $f_\theta$ should predict zero as a signal-independent term as follows:
\begin{equation}\label{eq:input_noise_w}
f_{\theta} \parens{ \hat{I}_{c} + \hat{I}_{c}^{\gamma}\hat{N}_{d} }
= f_{\theta} \parens{ \hat{I}_n^\text{dep} }
= \bracket{ \hat{I}_c, \hat{N}_{d}, \hat{\mathbf{0}} },
\end{equation}
where $\hat{I}_n^\text{dep}$ is a predicted signal-dependent noisy image.

For a given pure signal-independent part $\hat{N}_i$, we can regard that the corresponding clean image part is zero.
In other word, we can rewrite $\hat{N}_i = g ( \mathbf{0}, \ast, \hat{N}_i )$, where $s_1 = 0$ and $s_3 = 1$.
Here, $\paren{ \ast }$ denotes that we do not care about the signal-dependent part.
Therefore, our $f_\theta$ should predict the same noise $\hat{N}_{i}$ for the signal-independent path, and also zero for the clean image branch as follows:
\begin{equation}
    f_{\theta} \parens{ \hat{N}_{i} } = \bracket{ \hat{\mathbf{0}}, \ast, \hat{N}_{i} },
    \label{eq:input_noise_b}
\end{equation}
%
where we cannot identify the signal-dependent part $\paren{ \ast }$ as the predicted clean image should be zero.

Lastly, we simulate virtual synthetic noisy images
by combination of predicted outputs $\hat{I}_c$, $\hat{N}_d$, and $\hat{N}_i$ with the various scalar factors $(s_1,s_2,s_3)$ as shown in Figure~\ref{fig:framework_b}.
Therefore, we generate the augmented inputs by setting $s_1=1$ and selecting $s_2$ and $s_3$ from $\{-1,0,1\}$ and apply the $f_{\theta}$ as follows:
\begin{equation}
    f_{\theta} \parens{ \hat{I}_n^\text{aug} } = \bracket{ \hat{I}_{c}, s_2\hat{N}_{d}, s_3\hat{N}_{i}},
    \label{eq:ss_augmentation}
\end{equation}
where $\hat{I}_n^\text{aug} = \hat{I}_{c} + s_2\hat{I}_{c}^{\gamma} \hat{N}_{d} + s_3\hat{N}_{i}$.
This approach operates like self-supervised data augmentation, where no extra samples are required.

\subsection{Network architecture}
We employ a CNN model with sequential layers as the learnable $f_{\theta}$ as shown in Figure~\ref{fig:network}.
Our model consists of two parts: the clean image generator and the noise generator.
Given a noisy image, we employ DnCNN~\cite{DBLP:journals/corr/ZhangZCM016} without the skip connection and batch normalization layers (BN) as the clean generator which aims to reconstruct the corresponding noise-free output $\hat{I}_c$.
Then, we subtract the output from the noisy input image and feed this noise signal, \ie, $I_n - \hat{I}_c$, to sequential convolutional layers with two branches to estimate $N_{d}$ and $N_{i}$, respectively. We provide more details in the supplementary material.

\subsection{Loss functions for self-supervised learning}
\label{ssec:loss}
To train our network $f_{\theta}$, we define a set of loss functions based on statistical behaviors of general noise.
First, we define the consistency loss $\mathcal{L}^\text{con}$ to ensure the combination $g$ of outputs $\hat{I}_c$, $\hat{N}_{d}$ and $\hat{N}_{i}$ converges to noisy input as follows:
\begin{equation}\label{eq:consistency_loss}
\mathcal{L}^\text{con} = \normtwo{ I_n - g \paren{ f_{\theta} \paren{ I_n } } }.
\end{equation}
For simplicity, we use $\normtwo{ \cdot }$ to represent the $L^2$ norm.
%
\begin{table*}[t]
    \small
    \centering
    \begin{tabularx}{\linewidth}{c c l >{\centering\arraybackslash}X >{\centering\arraybackslash}X >{\centering\arraybackslash}X >{\centering\arraybackslash}X}
    \toprule
    \multirow{2}{*}{\textbf{Type of supervision}} & \multirow{2}{*}{\textbf{Training data}} & \multirow{2}{*}{\bf Method} & \multicolumn{2}{c}{\textbf{SIDD}} & \multicolumn{2}{c}{\textbf{DND}}\\
    & & & \textbf{PSNR} & \textbf{SSIM} & \textbf{PSNR} & \textbf{SSIM} \\
    \midrule
    \multirow{7}{*}{Supervised} &\multirow{6}{*}{Paired noisy/clean}
    & MLP~\cite{burger2012image} & {24.71} & {0.641} & {34.23} & {0.833} \\
    & & TNRD~\cite{7527621}& {24.73} & {0.643} & {33.65} & {0.830}  \\
    & & DnCNN~\cite{DBLP:journals/corr/ZhangZCM016} & {23.66} & {0.583} & {32.43} & {0.790}\\
    & & DnCNN+~\cite{DBLP:journals/corr/ZhangZCM016} & {32.59} & {0.861} & {37.90} & {0.943}\\
    & & CBDNet~\cite{DBLP:journals/corr/abs-1807-04686} & {33.28} & {0.868} & {38.05} & {0.942} \\
    & & RIDNet~\cite{DBLP:journals/corr/abs-1904-07396}  & {38.70} & {0.950} & {39.25} & {0.952} \\
    & & DIDN~\cite{9025411}  & {\textbf{39.82}} & {\textbf{0.973}} & {\textbf{39.62}} & {\textbf{0.954}} \\
    \midrule
    \multirow{4}{*}{Unsupervised} & \multirow{3}{*}{Unpaired noisy/clean}
    & GCBD~\cite{Chen2018ImageBD}  & {-} & {-} & {35.58} & {\textbf{0.922}} \\
    & & UIDNet~\cite{Hong_Fan_Jiang_Feng_2020} &  {32.48} & {0.897} & {-} & {-} \\
    & & C2N~\cite{Jang_2021_ICCV}  & {\textbf{35.35}} & {\textbf{0.937}} & {\textbf{36.38}} & {0.887} \\
    \cline{2-7}
    & \multirow{1}{*}{Paired noisy/noisy}
    & R2R~\cite{Pang_2021_CVPR} &  34.78 & {0.844} & {-} & {-} \\
    \midrule
    \multirow{4}{*}{Self-supervised} & \multirow{3}{*}{Paired noisy/noisy}
     & N2V~\cite{DBLP:journals/corr/abs-1811-10980}  & {27.68} & {0.668} & {-} & {-} \\
    & & N2S~\cite{DBLP:journals/corr/abs-1901-11365}  & {29.56} & {0.808}  & {-} & {-}\\
    & & NAC~\cite{NAC}  & {-} & {-}  & {36.20} & {\textbf{0.925}}\\
    \cline{2-7}
    & \multirow{3}{*}{Single noisy}
    & CVF-SID (T) (Ours) & 34.43 & 0.912 & 36.31 & 0.923 \\
    & & CVF-SID (S) (Ours) & 34.51 & 0.916 & 36.49 & 0.924 \\  
    & & \textbf{CVF-SID} ($\textbf{S}^2$) (\textbf{Ours}) & \textbf{34.71} & \textbf{0.917} & \textbf{36.50} & 0.924\\    
    \bottomrule
    \end{tabularx}
    \vspace{-2mm}
    \caption{
        \textbf{Quantitative comparison of real-world  sRGB image denoising on SIDD and DND benchmark datasets.}
        We compare CVF-SID with other denoising methods in terms of PSNR and SSIM.
        \textbf{T}, \textbf{S}, and $\textbf{S}^2$ refer to different training strategies discussed in Section~\ref{4.3}.
    }
    \label{tab:SIDD_benchmark}
    \vspace{-4mm}
\end{table*}

Moreover, we construct the identity loss $\mathcal{L}^\text{id}$ based on our inter-dependency assumption as follows:
\begin{equation}
    \begin{split}
        \mathcal{L}^\text{id} &= \normtwo{ \hat{I}_c - f_{\theta}^\text{clean} \parens{ \hat{I}_c } } + \normtwo{ \hat{I}_c - f_{\theta}^\text{clean} \parens{ \hat{I}_n^\text{dep} } } \\
        &+ \normtwo{ \hat{N}_d - f_{\theta}^\text{dep} \parens{ \hat{I}_n^\text{dep} } } + \normtwo{ \hat{N}_i - f_{\theta}^\text{indep} \parens{ \hat{N}_i } }.
    \end{split}
    \label{eq:identity_loss}
\end{equation}
When our network takes a denoised image $\hat{I}_c$ as an input, we expect the model to predict the same image as output without any noise terms.
We also define similar loss training objectives for images corrupted with pure signal-dependent noise $\hat{I}_n^\text{dep}$ and the predicted signal-independent noise $\hat{N}_i$ to construct our identity loss $\mathcal{L}^\text{id}$.
%

%
On the other hand, we design the zero loss $\mathcal{L}^\text{zero}$ to satisfy the constraints in Eq.~\ref{eq:input_clean}, Eq.~\ref{eq:input_noise_w}, and Eq.~\ref{eq:input_noise_b} as follows:
\begin{equation}\label{eq:zero_loss}
\begin{split}
    \mathcal{L}^\text{zero}
    &= \normtwo{ f_{\theta}^\text{dep} \parens{ \hat{I}_c} }
    + \normtwo{ f_{\theta}^\text{indep} \parens{ \hat{I}_c} } \\
    &+ \normtwo{f_{\theta}^\text{clean} \parens{ \hat{N}_i}}
    + \normtwo{f_{\theta}^\text{indep} \parens{ \hat{I}_n^\text{dep} }}.
\end{split}
\end{equation}

While Eq.~\ref{eq:consistency_loss}, Eq.~\ref{eq:identity_loss}, and Eq.~\ref{eq:zero_loss} provide several constraints for our self-supervised framework, we further introduce a regularization term to avoid trivial solutions, \eg, zero-valued noise.
Our assumption is that the variance of the predicted noise should be positive.
Inspired by the patch-based local estimation~\cite{Liu2014PracticalSN}, we first approximate the noise variance from the given image by averaging the variances of $M$ small patches $I_{n}^{j}$ as follows:
\begin{equation}\label{eq:variance}
    \begin{split}
        \mathrm{Var} \parens{ I_{n}^{j} }
        &\approx \mathrm{Var} \parens{ \hat{I}^j_c + \hat{I}_c^{j\gamma} \hat{N}_{d}^j } + \mathrm{Var} \parens{ \hat{N}_{i}^j }\\
        &= C_j^{2\gamma} \hat{\sigma}^2_{j, d} + \hat{\sigma}^2_{j, i} = C_j^{2\gamma} \hat{\sigma}^2_{d} + \hat{\sigma}^2_{i},
    \end{split}
\end{equation}
where $\hat{\sigma}_{j, \ast}^2 = \mathrm{Var} \parens{\hat{N}_\ast^j}$ for $\ast = d$ or $i$.
We assume that an ideal clean image is approximately constant for a $j$-th patch, \ie, $I^j_c \approx C_j$ and $\mathrm{Var} \parens{ I^j_c} \approx 0$, for a \emph{small local region}.
Also, we assume that $N_{d}$ and $N_{i}$ are spatially uncorrelated, which means that $\hat{\sigma}_{j, \ast}^2 = \hat{\sigma}_{ \ast}^2 $ is a constant in the image space.
Accordingly, we define the regularization loss $\mathcal{L}^\text{reg}$ to prevent the estimated noise map from having zero-variance, \ie, trivial solution, as follows:
\begin{equation}\label{eq:variance_loss}
\begin{split}
    \mathcal{L}^\text{reg}
    &=\frac{1}{M} { \normtwo{ \sum_{j = 1}^M \mathrm{Var}(I_{n}^{j}) - \sum_{j = 1}^M C_j^{2\gamma} \hat{\sigma}^2_{d} - M\hat{\sigma}^2_{i}} }.
\end{split}
\end{equation}
Our regularization loss $\mathcal{L}^\text{reg}$ in Eq.~\ref{eq:variance_loss} is constructed based on an assumption that outputs of the network $f_\theta$ are independent to each other.
%
%

%
Finally, we define the augmentation loss $\mathcal{L}^\text{aug}$ for each of the augmented data in Eq.~\ref{eq:ss_augmentation} as follows:
\begin{equation}
    \begin{split}
        \mathcal{L}^\text{aug}
        &= \normtwo{ f_\theta^\text{clean} \parens{ \hat{I}_n^\text{aug} } - \hat{I}_c } 
        + \normtwo{ f_\theta^\text{dep} \parens{ \hat{I}_n^\text{aug} } - s_2\hat{N}_d }\\
        & + \normtwo{ f_\theta^\text{indep} \parens{ \hat{I}_n^\text{aug} } - s_3\hat{N}_i }, \\
    \end{split}
    \label{eq:loss_aug}
\end{equation}
where we calculate Eq.~\ref{eq:loss_aug} for all possible combinations of $s_2$ and $s_3$ in the augmented sample $\hat{I}_N^\text{aug}$.
Our total training objective $\mathcal{L}^\text{total}$ is defined by the summation of all aforementioned loss functions as follows:
\begin{equation}
    \mathcal{L}^\text{total} = \mathcal{L}^\text{con} + \mathcal{L}^\text{id} + \mathcal{L}^\text{zero} + \mathcal{L}^\text{reg} +  \lambda_{\text{aug}}\mathcal{L}^\text{aug}.
    \label{eq:loss_total}
\end{equation}


%

\section{Experiments}
In this section, we first discuss the datasets as well as detailed configurations used to train our CVF-SID framework.
We then describe comprehensive experimental results and extensive comparisons with the other methods.
\begin{figure*}
    \centering
    \captionsetup[subfigure]{labelformat=empty}
    \renewcommand{\wp}{0.119\linewidth}
    \subfloat{\includegraphics[width=\wp,
    page=1]{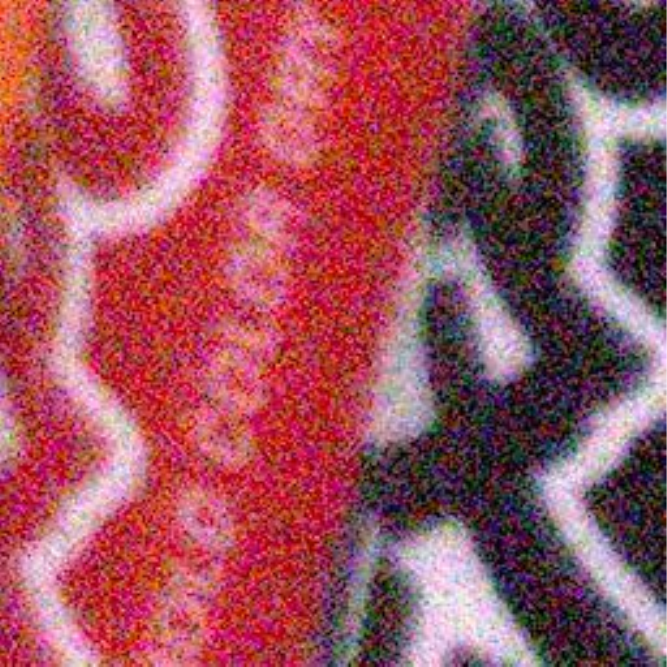}}
    \hfill
    \subfloat{\includegraphics[width=\wp, page=2]{figures/image_310.pdf}}
    \hfill
    \subfloat{\includegraphics[width=\wp, page=3]{figures/image_310.pdf}}
    \hfill
    \subfloat{\includegraphics[width=\wp, page=4]{figures/image_310.pdf}}
    \hfill
    \subfloat{\includegraphics[width=\wp, page=5]{figures/image_310.pdf}}
    \hfill
    \subfloat{\includegraphics[width=\wp, page=6]{figures/image_310.pdf}}
    \hfill
    \subfloat{\includegraphics[width=\wp, page=7]{figures/image_310.pdf}}
    \hfill
    \subfloat{\includegraphics[width=\wp, page=9]{figures/image_310.pdf}}
    \\
    \subfloat{\includegraphics[width=\wp, page=1]{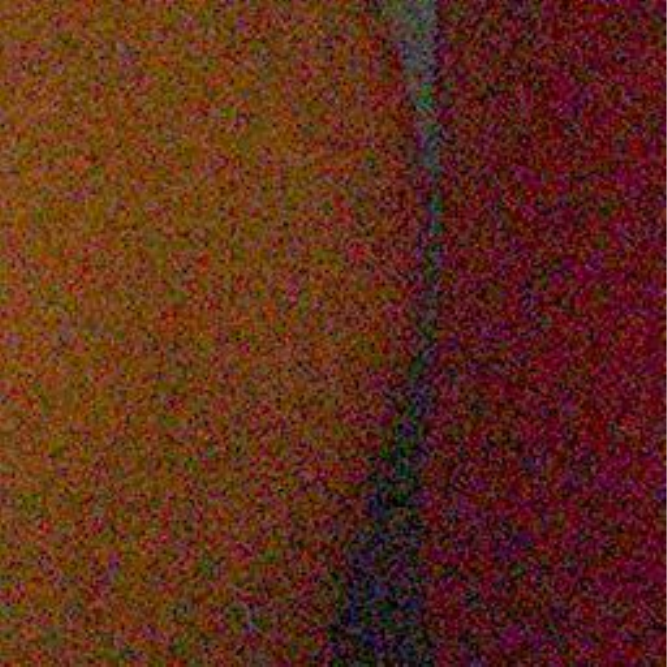}}
    \hfill
    \subfloat{\includegraphics[width=\wp, page=2]{figures/image_704.pdf}}
    \hfill
    \subfloat{\includegraphics[width=\wp, page=3]{figures/image_704.pdf}}
    \hfill
    \subfloat{\includegraphics[width=\wp, page=4]{figures/image_704.pdf}}
    \hfill
    \subfloat{\includegraphics[width=\wp, page=5]{figures/image_704.pdf}}
    \hfill
    \subfloat{\includegraphics[width=\wp, page=6]{figures/image_704.pdf}}
    \hfill
    \subfloat{\includegraphics[width=\wp, page=7]{figures/image_704.pdf}}
    \hfill
    \subfloat{\includegraphics[width=\wp, page=9]{figures/image_704.pdf}}
    \\
    \subfloat{\includegraphics[width=\wp, page=1]{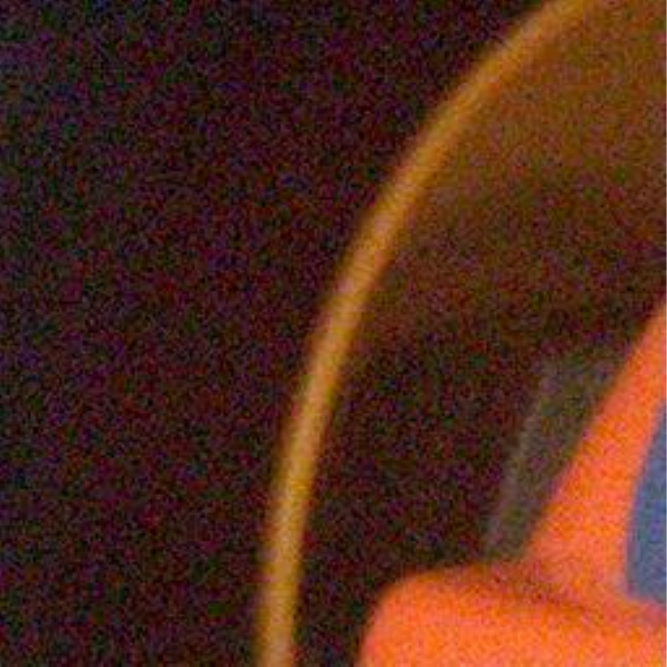}}
    \hfill
    \subfloat{\includegraphics[width=\wp, page=2]{figures/image_852.pdf}}
    \hfill
    \subfloat{\includegraphics[width=\wp, page=3]{figures/image_852.pdf}}
    \hfill
    \subfloat{\includegraphics[width=\wp, page=4]{figures/image_852.pdf}}
    \hfill
    \subfloat{\includegraphics[width=\wp, page=5]{figures/image_852.pdf}}
    \hfill
    \subfloat{\includegraphics[width=\wp, page=6]{figures/image_852.pdf}}
    \hfill
    \subfloat{\includegraphics[width=\wp, page=7]{figures/image_852.pdf}}
    \hfill
    \subfloat{\includegraphics[width=\wp, page=9]{figures/image_852.pdf}}
    \\
    \addtocounter{subfigure}{-25}
    \subfloat[\scriptsize Input Noisy]{\includegraphics[width=\wp, page=1]{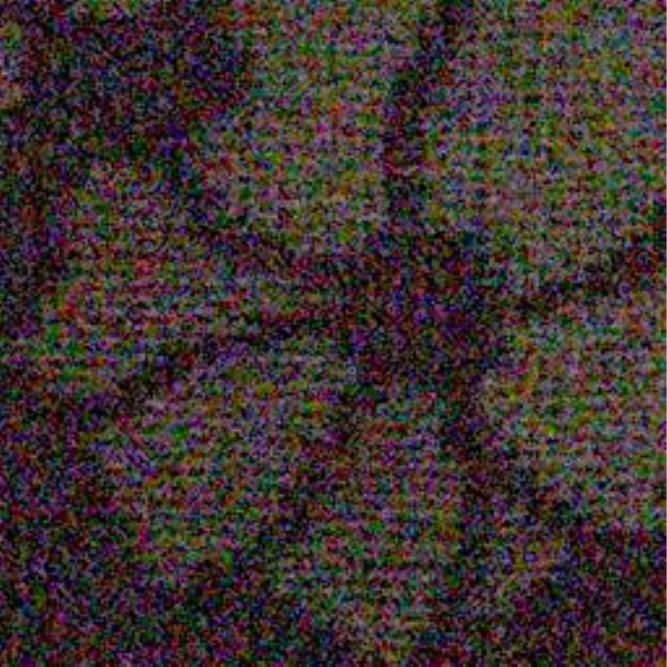}}
    \hfill
    \subfloat[\scriptsize BM3D~\cite{sparse}]{\includegraphics[width=\wp, page=2]{figures/image_1140.pdf}}
    \hfill
    \subfloat[\scriptsize NC~\cite{ipol.2015.125}]{\includegraphics[width=\wp, page=3]{figures/image_1140.pdf}}
    \hfill
    \subfloat[\scriptsize N2V~\cite{DBLP:journals/corr/abs-1811-10980}]{\includegraphics[width=\wp, page=4]{figures/image_1140.pdf}}
    \hfill
    \subfloat[\scriptsize N2S~\cite{DBLP:journals/corr/abs-1901-11365}]{\includegraphics[width=\wp, page=5]{figures/image_1140.pdf}}
    \hfill
    \subfloat[\scriptsize DnCNN~\cite{DBLP:journals/corr/ZhangZCM016}]{\includegraphics[width=\wp, page=6]{figures/image_1140.pdf}}
    \hfill
    \subfloat[\scriptsize R2R~\cite{Pang_2021_CVPR}]{\includegraphics[width=\wp, page=7]{figures/image_1140.pdf}}
    \hfill
    \subfloat[\scriptsize \textbf{CVF-SID (Ours)}]{\includegraphics[width=\wp, page=9]{figures/image_1140.pdf}}
    \\
    \vspace{-3mm}
    \caption{
        \textbf{Qualitative comparison of different denoising methods on SIDD benchmark.} 
    }
    \label{fig:SIDD_benchmark}
    \vspace{-3mm}
\end{figure*}
\begin{figure*}[t]
	\captionsetup[subfloat]{labelformat=empty}
		\newcommand{\rowArg}{1.5cm}
		\newcommand{\fullSize}{4.5cm}
		\newcommand{\patchSize}{1.8cm}
		\setlength\tabcolsep{0.05cm}
		\begin{tabular}[b]{c c c}
			\multicolumn{2}{c}{\multirow{2}{*}[\rowArg]{
					\subfloat[\scriptsize Input Noisy]
					{\includegraphics[page=1, height=\fullSize, trim={0.9cm 0 0.9cm 0}, clip]
						{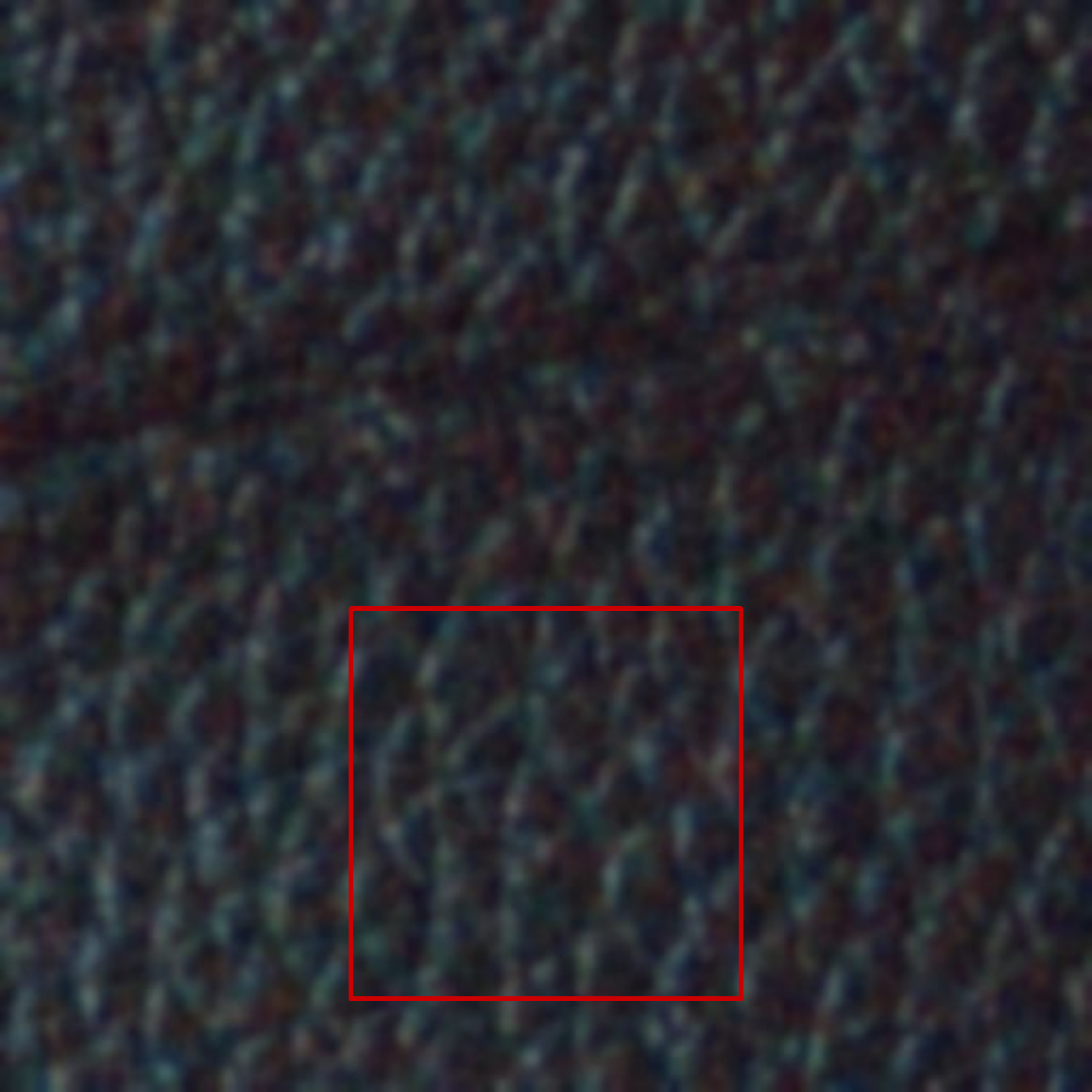}}}} &
			\subfloat[\scriptsize \centering DnCNN+~\cite{DBLP:journals/corr/ZhangZCM016} \protect\linebreak 31.34dB / 0.834 ]{
				\includegraphics[page=5, width = \patchSize, height = \patchSize]
				{figures/DND_308.pdf}} 
				\vspace{2.25mm}
			\subfloat[\scriptsize \centering C2N~\cite{Jang_2021_ICCV}  \protect\linebreak 28.37dB / 0.629]{
				\includegraphics[page=3, width = \patchSize, height = \patchSize]
				{figures/DND_308.pdf}} \\  & &

			\subfloat[\scriptsize \centering NAC~\cite{NAC}  \protect\linebreak 31.23dB / \textbf{0.853} ]{
				\includegraphics[page=9, width = \patchSize, height = \patchSize]
				{figures/DND_308.pdf}}
				\vspace{2.25mm}
			\subfloat[\scriptsize \centering \textbf{CVF-SID (Ours)}   \protect\linebreak \textbf{31.84dB} / 0.832 ]{
				\includegraphics[page=2, width = \patchSize, height = \patchSize]
				{figures/DND_308.pdf}}
	\end{tabular}
	\hfill
		\setlength\tabcolsep{0.05cm}
		\begin{tabular}[b]{c c c}
			\multicolumn{2}{c}{\multirow{2}{*}[\rowArg]{
					\subfloat[\scriptsize Input Noisy]
					{\includegraphics[page=1, height=\fullSize, trim={0.9cm 0 0.9cm 0}, clip]
						{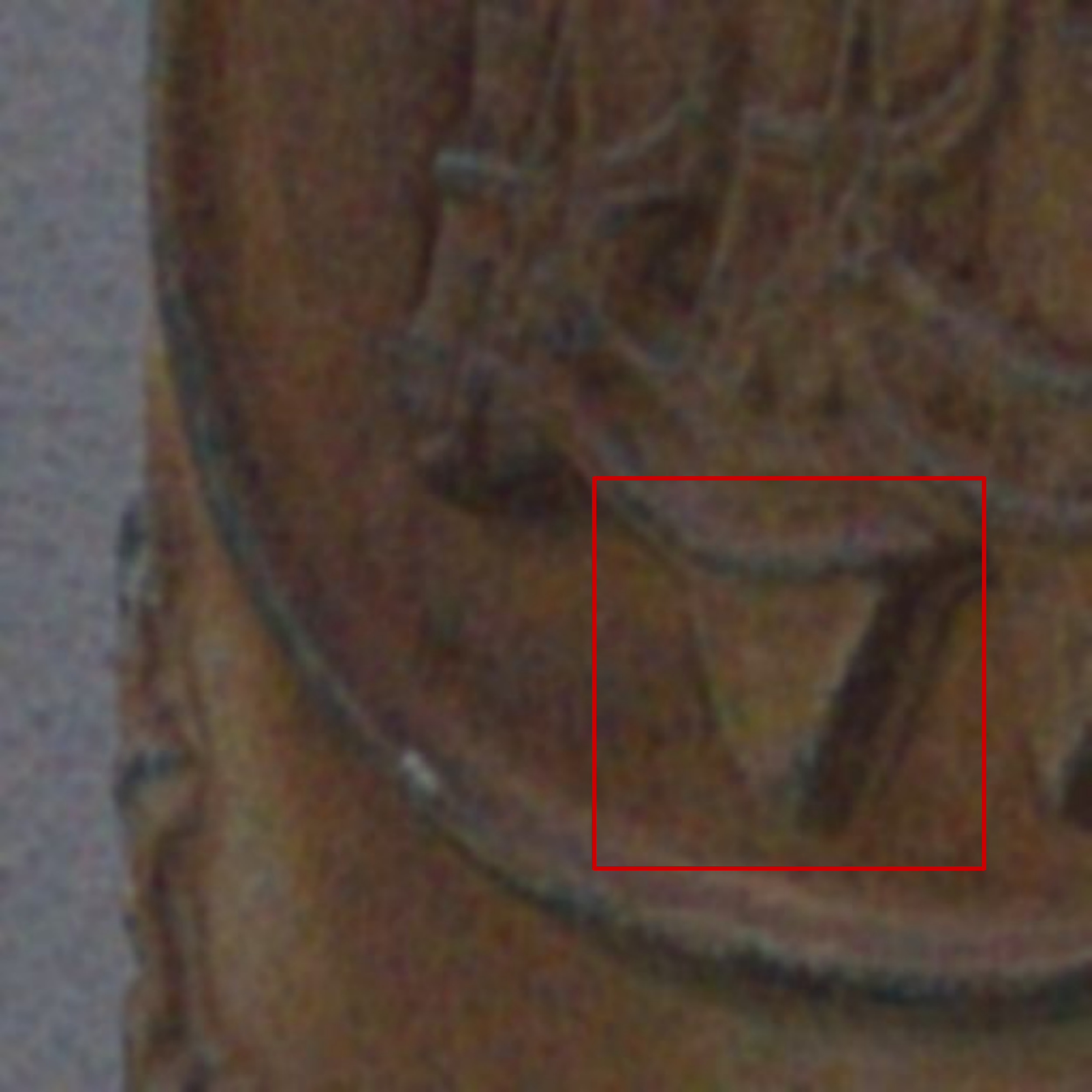}}}} &
			\subfloat[\scriptsize \centering DnCNN+~\cite{DBLP:journals/corr/ZhangZCM016}  \protect\linebreak 34.51dB / 0.946 ]{
				\includegraphics[page=5, width = \patchSize, height = \patchSize]
				{figures/DND_501.pdf}} 
				\vspace{2.25mm}
			\subfloat[\scriptsize \centering C2N~\cite{Jang_2021_ICCV}  \protect\linebreak 33.67dB / 0.927 ]{
				\includegraphics[page=3, width = \patchSize, height = \patchSize]
				{figures/DND_501.pdf}} \\  & &

			\subfloat[\scriptsize \centering NAC~\cite{NAC}  \protect\linebreak 34.21dB / 0.934 ]{
				\includegraphics[page=9, width = \patchSize, height = \patchSize]
				{figures/DND_501.pdf}}
				\vspace{2.25mm}
			\subfloat[\scriptsize \centering \textbf{CVF-SID (Ours)} \protect\linebreak \textbf{34.74dB} / \textbf{0.951} ]{
				\includegraphics[page=2, width = \patchSize, height = \patchSize]
				{figures/DND_501.pdf}}
				\end{tabular}
	\setlength{\abovecaptionskip}{0cm}
	\vspace{-2mm}
	\caption{\textbf{Visual comparison of different denoising methods on DND benchmark.} 
	    DnCNN+~\cite{DBLP:journals/corr/ZhangZCM016}, C2N~\cite{Jang_2021_ICCV}, and NAC~\cite{NAC} are supervised, unsupervised, and self-supervised methods, respectively.
	    We report PSNR/SSIM of each result \wrt clean ground-truth image.
	}
	\label{fig:DND_benchmark}
	\vspace{-4mm}
\end{figure*}

\subsection{Dataset}\label{4.1}
We train and evaluate our method on two datasets, including real-world noisy images: SIDD~\cite{8578280} and  DND~\cite{DBLP:journals/corr/PlotzR17}.
%
%
\Paragraph{Smartphone Image Denoising Dataset (SIDD)}~\cite{8578280} is one of the representative real-world datasets which contains well-aligned noisy-clean image pairs for training.
We use sRGB images from the SIDD-Medium dataset for training, including $320$ noisy-clean image pairs.
For evaluation, validation and benchmark splits are used that each contains 32 patches of size $256 \times 256$ from 40 images where no ground-truth clean images are provided for the benchmark track.
%

%
\Paragraph{Darmstadt Noise Dataset (DND)}~\cite{DBLP:journals/corr/PlotzR17} benchmark consists of $50$ noisy images captured with consumer-grade cameras of various sensor sizes.
Each image is cropped into $20$ patches of size $512\times 512$, where total $50 \times 20 = 1,000$ samples are provided for evaluation.
Compared to the SIDD dataset, images in the DND dataset are captured under normal lighting conditions and therefore contain weaker noise.

%

%

\subsection{Training details}\label{4.2}
During the training, we construct a mini-batch of size
$64$, which contains $40 \times 40$ random crops of training images.
Random flip and rotation augmentation is applied to increase the number of effective training samples.
The learning rate is set to $10^{-4}$, where ADAM~\cite{kingma2017adam} optimizer is used to update the learnable parameters.
For all of our experiments, we fix $\gamma = 1$ in Eq.~\ref{eq:g_function}.
To apply the regression loss $\mathcal{L}^\text{reg}$ in Eq.~\ref{eq:variance_loss}, we densely extract $6 \times 6$ patches for each pixel in input images to calculate the approximated variance.
In Eq.~\ref{eq:loss_total}, we set $\lambda_\text{aug} = 0.1$ to calculate the total loss $\mathcal{L}^\text{total}$.
We adopt peak signal-to-noise ratio (PSNR) and structural similarity (SSIM) as evaluation metrics for quantitative comparison.
All experiments are done using PyTorch 1.7.1 and Quadro RTX 8000 GPUs.
We note that the inference time is about 10ms on average for a given $256\times256$ input image with only the clean image generator.

\subsection{Evaluation on real-world sRGB datasets}
\label{4.3}
We evaluate our CVF-SID on real-world sRGB noisy images from SIDD validation, SIDD benchmark, and DND benchmark datasets.
For the SIDD and DND benchmarks, we submit the denoising results to websites for server-side evaluation.
To evaluate our model on the three different datasets, we leverage three different training strategies.
We refer each of them to \textbf{T}, \textbf{S}, and $\textbf{S}^2$, respectively.

\Paragraph{Training on SIDD-Medium dataset (T).}
In the first scenario, we train our CVF-SID on noisy images from the SIDD-Medium dataset.
The learned model is then evaluated on three datasets.

\Paragraph{Training on test dataset (S).}
Since CVF-SID is fully self-supervised, we can train our model on the same dataset used for evaluation.
Therefore, we train our method on three evaluation datasets, respectively, and test each of them using the same dataset.
Since the noise distribution of training and test samples are matched, such a strategy leads to better adaptation to the evaluation datasets.

\Paragraph{Double-denoising on test dataset ($\text{S}^2$}).
After training on the test dataset (S), we apply our learned CVF-SID to the test images to acquire denoised images.
Then, we use the denoised images as a new dataset to double-train the second CVF-SID.
The final denoised results are restored by two successive CVF-SID models on the original noisy images.

Table~\ref{tab:SIDD_benchmark} shows extensive comparisons between several supervised, un-/self-supervised denoising methods on SIDD and DND benchmarks.
We note that (T), (S), and ($\text{S}^2$) denote our different training strategies described above, respectively.
Interestingly, CVF-SID (S) slightly outperforms CVF-SID (T), which is trained on a large SIDD-Medium dataset.
This observation validates the advantage of our method, which can be directly trained on test sRGB images without requiring Raw-RGB data or a large number of training samples.
Moreover, evaluation of CVF-SID (T) on DND (trained on SIDD, tested on DND) shows the generalization ability of CVF-SID for out-of-domain (or cross-domain) image denoising.
Furthermore, we show that CVF-SID can be trained in a recursive fashion ($\text{S}^2$) on the actual test dataset to achieve better denoising performance.
Without using any clean images, CVF-SID ($\text{S}^2$) outperforms several un-/self-supervised methods on both SIDD and DND benchmarks.
Figure~\ref{fig:SIDD_benchmark} and \ref{fig:DND_benchmark} show qualitative comparisons between different denoising methods on these datasets.

The first row of Figure~\ref{fig:SIDD_benchmark} shows that N2S and DnCNN cannot reconstruct characters, while BM3D and N2V cannot perfectly remove noise.
In contrast, our CVF-SID can remove unpleasant noise while preserving text details.
In the fourth row, we can see that NC and R2R cannot preserve detailed textures while ours can.
On the left side of Figure~\ref{fig:DND_benchmark}, the proposed CVF-SID can reconstruct detailed textures while removing the noise.
We note that NAC cannot preserve the original colors compared to the other methods.
On the right side, our CVF-SID can preserve edges while suppressing noise from the input.
%


We also evaluate CVF-SID on SIDD validation dataset using three different training strategies as shown in Table~\ref{tab:SIDD_validation}.
Our approach performs much better than the existing self-supervised methods and unsupervised method C2N~\cite{Jang_2021_ICCV} and even achieves  comparable performance to recent unsupervised R2R~\cite{Pang_2021_CVPR}.
A major merit of CVF-SID compared to R2R~\cite{Pang_2021_CVPR} and other self-supervised approaches is that we do not generate any paired auxiliary noisy images.
Also, we do not assume any specific distribution regarding the unknown noise signals, making our method more generalizable.

\begin{figure}
    \centering
    \captionsetup[subfigure]{labelformat=empty}
    \renewcommand{\wp}{0.190 \linewidth}
    \begin{minipage}[t]{0.99 \linewidth}
        \subfloat{\includegraphics[width=\wp, page=1]{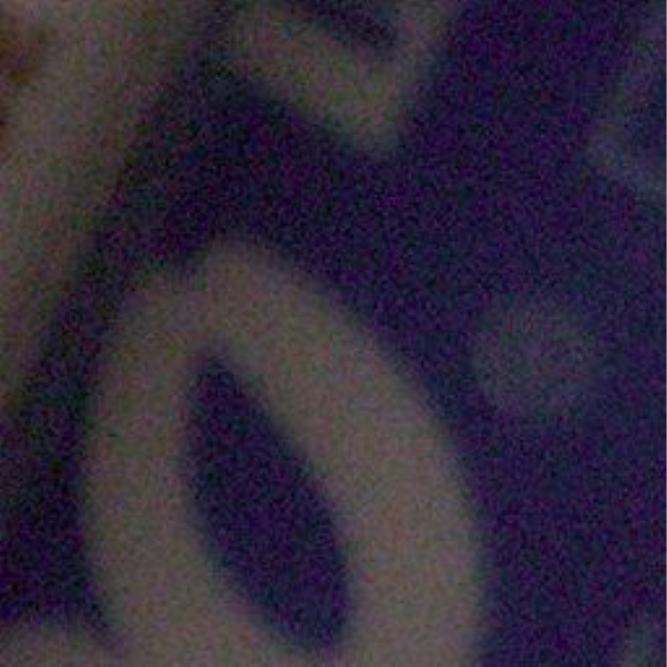}}
        \hfill
        \subfloat{\includegraphics[width=\wp, page=2]{figures/image_252.pdf}}
        \hfill
        \subfloat{\includegraphics[width=\wp, page=3]{figures/image_252.pdf}}
        \hfill
        \subfloat{\includegraphics[width=\wp, page=4]{figures/image_252.pdf}}
        \hfill
        \subfloat{\includegraphics[width=\wp, page=5]{figures/image_252.pdf}}
        \\
        \subfloat{\includegraphics[width=\wp, page=1]{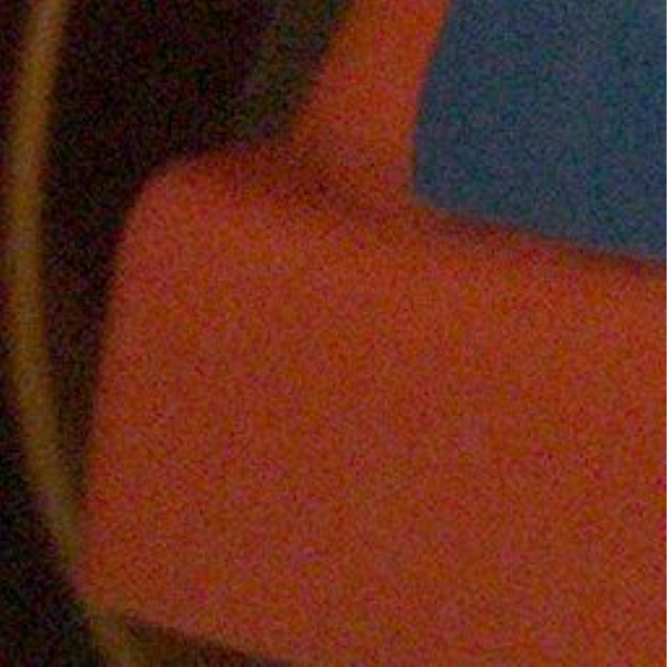}}
        \hfill
        \subfloat{\includegraphics[width=\wp, page=2]{figures/image_803.pdf}}
        \hfill
        \subfloat{\includegraphics[width=\wp, page=3]{figures/image_803.pdf}}
        \hfill
        \subfloat{\includegraphics[width=\wp, page=4]{figures/image_803.pdf}}
        \hfill
        \subfloat{\includegraphics[width=\wp, page=5]{figures/image_803.pdf}}\\
        \subfloat{\includegraphics[width=\wp, page=1]{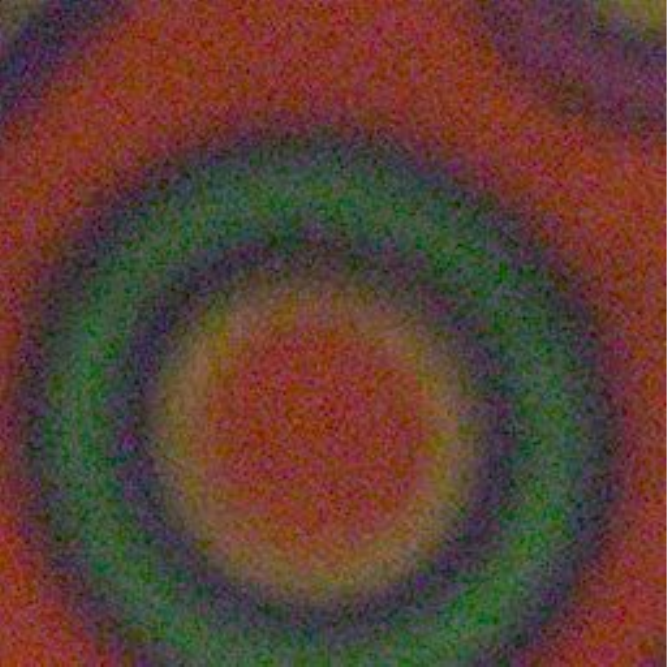}}
        \hfill
        \subfloat{\includegraphics[width=\wp, page=2]{figures/image_543.pdf}}
        \hfill
        \subfloat{\includegraphics[width=\wp, page=3]{figures/image_543.pdf}}
        \hfill
        \subfloat{\includegraphics[width=\wp, page=4]{figures/image_543.pdf}}
        \hfill
        \subfloat{\includegraphics[width=\wp, page=5]{figures/image_543.pdf}}
        \\
        \subfloat[\scriptsize $I_n$]{\includegraphics[width=\wp, page=1]{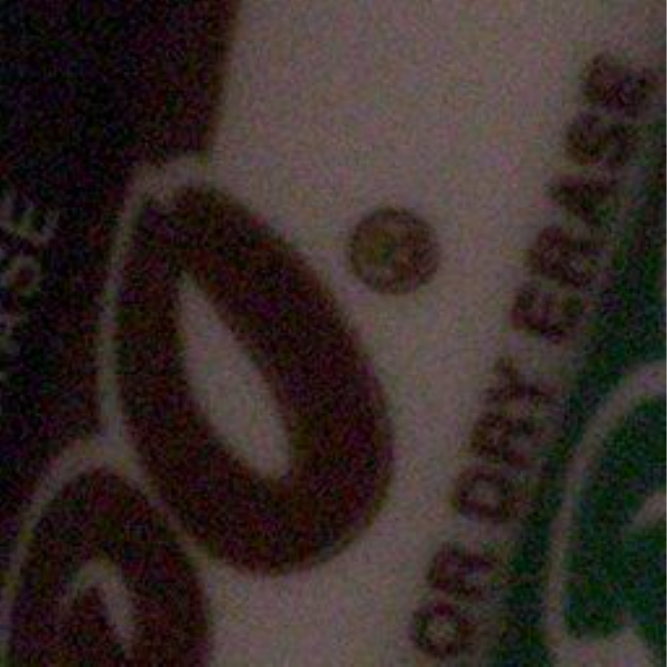}}
        \hfill
        \subfloat[\scriptsize $I_c$]{\includegraphics[width=\wp, page=2]{figures/image_246.pdf}}
        \hfill
        \subfloat[\scriptsize $\hat{I}_c$]{\includegraphics[width=\wp, page=3]{figures/image_246.pdf}}
        \hfill
        \subfloat[\scriptsize $\hat{N}_d$]{\includegraphics[width=\wp, page=4]{figures/image_246.pdf}}
        \hfill
        \subfloat[\scriptsize $\hat{N}_i$]{\includegraphics[width=\wp, page=5]{figures/image_246.pdf}}
    \end{minipage}
    \vspace{-2mm}
    \caption{
        \textbf{Decomposition results of our CVF-SID (S) on the SIDD validation dataset.}
        %
        %
        %
        For better visualization, the noise maps $\hat{N}_d$ and $\hat{N}_i$ are normalized to $[ 0, 1 ]$.
    }
    \label{fig:decomposition}
    \vspace{-4mm}
\end{figure}

As a result, our approach can be applied to sRGB images directly while R2R requires Raw-RGB images for pre-training.
We note that Raw-RGB color space contains more information than sRGB, and thus using Raw-RGB samples usually yields better performance than the pure sRGB configuration~\cite{8578280, DBLP:journals/corr/PlotzR17, abdelhamed2019ntire}.
Since most digital images are stored in the sRGB format, our CVF-SID can handle more general inputs than R2R.
%
In Figure~\ref{fig:decomposition}, we visualize how CVF-SID decomposes the given noisy image into the clean image, signal-dependent, and signal-independent noises.
We attach more visual comparisons in the supplementary material.
\begin{table}[t]
    \small
    \centering
    \setlength\tabcolsep{2pt} 
    \begin{tabularx}{\linewidth}{c l >{\centering\arraybackslash}X >{\centering\arraybackslash}X}
        \toprule
        & \textbf{Method} & \textbf{PSNR} & \textbf{SSIM} \\
        \midrule
        \multirow{4}{*}{Non-learning based}  
        & BM3D~\cite{sparse} & {25.65} & {0.475}  \\
        & WNNM~\cite{6909762}& {26.20} & {0.693}\\
        & NC~\cite{ipol.2015.125} & {31.31} & {0.725}  \\
        & MCWNNM~\cite{DBLP:journals/corr/XuZZF17}& {\textbf{33.40}} & {\textbf{0.815}}\\
        \midrule
        \multirow{2}{*}{Unsupervised}
        & C2N~\cite{Jang_2021_ICCV} & {34.08} & {-}  \\
        & R2R~\cite{Pang_2021_CVPR} & {\textbf{35.04}} & {\textbf{0.844}}  \\
        \midrule
        \multirow{5}{*}{Self-supervised}
        & N2V~\cite{DBLP:journals/corr/abs-1811-10980}& {29.35} & {0.651} \\
        & N2S~\cite{DBLP:journals/corr/abs-1901-11365} & {30.72} & {0.787} \\
        & CVF-SID (T) & 34.51 & 0.941 \\
        & CVF-SID (S)  & 34.67 & 0.943  \\
        & \textbf{CVF-SID} ($\textbf{S}^2$)  &  \textbf{34.81} & \textbf{0.944}  \\
        \bottomrule
    \end{tabularx}
    \vspace{-2mm}
    \caption{
        \textbf{Quantitative comparison of real-world denoising on sRGB images in SIDD validation dataset.}
    }
    \vspace{-4mm}
    \label{tab:SIDD_validation}
\end{table}


\subsection{Ablation study}\label{4.5}
In this section, we conduct some ablation studies to evaluate the performance of our proposed method better.

\Paragraph{Ablation on the loss function.}
As we have discussed in Section~\ref{ssec:loss}, we use various types of self-supervised loss terms to train our self-supervised CVF-SID.
Table~\ref{tab:loss} identifies the effect of each loss function for the training.
While the consistency term $\mathcal{L}^\text{con}$ is necessary to train our model, the identity loss $\mathcal{L}^\text{id}$ brings a significant performance gain and stabilize the learning process.

\Paragraph{Effect of the augmentation.}
To validate the effect of the proposed self-supervised augmentation strategies in Eq.~\ref{eq:loss_aug}, we also train our CVF-SID without and with the augmentation sets A and B in Figure~\ref{fig:framework_b}.
Table~\ref{tab:augmentation} shows the effect of each augmentation set, where set B brings about 0.2dB improvements in the denoising performance.
%

\begin{table}
    \centering
    \small
    \begin{tabularx}{\linewidth}{>{\centering\arraybackslash}X
    >{\centering\arraybackslash}X >{\centering\arraybackslash}X >{\centering\arraybackslash}X c c}
        \toprule
        \multicolumn{4}{c}{\textbf{Loss}} &  \multirow{2}{*}{\textbf{PSNR}} & \multirow{2}{*}{\textbf{SSIM}} \\
        $\mathcal{L}^\text{con}$ & $\mathcal{L}^\text{id}$ & $\mathcal{L}^\text{zero}$ & $\mathcal{L}^\text{reg}$ & & \\
        \midrule
        \ding{51} & $-$ & $-$ & $-$ & 33.10 & 0.923  \\
        \ding{51} & \ding{51} & $-$ & $-$ & 34.24 & \textbf{0.942}  \\
        \ding{51} & \ding{51} & \ding{51} & $-$ & 34.29 & 0.940 \\
        \ding{51} & \ding{51} & \ding{51} & \ding{51} & \textbf{34.43} & \textbf{0.942}  \\
        \bottomrule
    \end{tabularx}
    \\
    \vspace{-2mm}
    \caption{
        \textbf{Effect of loss terms for our CVF-SID (S) on the SIDD validation dataset.}
        We note that $\mathcal{L}^\text{aug}$ in Eq.~\ref{eq:loss_total} is not used.
        Please refer to Section~\ref{ssec:loss} for more details about each training objective.
        }
    \label{tab:loss}
    \vspace{-4mm}
\end{table} 
\begin{table}
    \centering
    \subfloat[\small Effect of the augmentations.]{
    \footnotesize
    \begin{tabularx}{0.48 \linewidth}{>{\centering\arraybackslash}X
    >{\centering\arraybackslash}X c c}
        \toprule
        \multicolumn{2}{c}{\textbf{Aug.}} &  \multirow{2}{*}{\textbf{PSNR}} &  \multirow{2}{*}{\textbf{SSIM}}  \\ 
        \textbf{A} & \textbf{B} & & \\
        \midrule
        $-$ & $-$ & 34.43 & 0.942  \\
        \ding{51} & $-$ & 34.48 & \textbf{0.943} \\
        \ding{51} & \ding{51} & \textbf{34.67} &  \textbf{0.943} \\
        \bottomrule
    \end{tabularx}\label{tab:augmentation}
    }
    \hfill
    \subfloat[\small Effect of the correlation $\gamma$.]{
    \footnotesize
    \begin{tabularx}{0.48 \linewidth}{>{\centering\arraybackslash}X c c}
        \toprule
        $\gamma$ & \textbf{PSNR} & \textbf{SSIM} \\
        \midrule
        0.25 & 34.45 & 0.942 \\
        0.50 & 34.46 & 0.942 \\
        \textbf{1.00} &  \textbf{34.67} & \textbf{0.943}  \\
        1.50 & 34.66 & \textbf{0.943} \\
        \bottomrule
    \end{tabularx}\label{tab:gamma}
    }
    \\
    \vspace{-2mm}
    \caption{
        \textbf{Effects of different hyperparameters for CVF-SID (S) on the SIDD validation dataset.}
        (a) An overview of the augmentation sets A and B is illustrated in Figure~\ref{fig:framework_b}.
        %
        (b) We find the best correlation parameter $\gamma$ by grid search.
    }
    \vspace{-4mm}
\end{table} 

\Paragraph{Ablation on $\gamma$ correlation.}
Following Torricelli~\etal~\cite{torricelli2002modelling}, we set the correlation parameter $\gamma$ in Eq.~\ref{eq:g_function} to $1$ to represent pure multiplicative noise.
Since real-world noise may exhibit more complex behavior, we conduct an ablation study regarding appropriate value for $\gamma$.
Table~\ref{tab:gamma} shows that CVF-SID achieves the best under the pure multiplicative assumption, \ie, $\gamma = 1$, while increasing the value does not change the performance much.

\section{Conclusion}

We propose CVF, a novel cyclic multi-variate function that decomposes an input under the cyclic procedure.
Then, we utilize CVF to design our self-supervised CVF-SID denoising framework, which aims to learn a CNN to disentangle the signal-dependent, signal-independent noises and clean image from a real-world noisy sRGB input.
The proposed approach does not rely on any prior information about the noise distribution, thus more generalizable than previous self-supervised denoising methods.
Extensive studies demonstrate several strengths and superiority of our formulation compared to the others.
One remaining limitation is that we resort to a fixed correlation parameter $\gamma$ in our framework, while the correlation may vary for different images in real-world applications.
This results in a sub-optimal decomposition, as shown in some examples of Figure~\ref{fig:decomposition}, where there exists little correlation between the image and signal-dependent noise term.
In our future work, we will also aim to learn the correlation parameter in a self-supervised manner while extending the concept of CVF toward various computer vision tasks.
%

\noindent
\textbf{Acknowledgement.}
This work was supported in part by IITP grant funded by the Korea government (MSIT) [No. 2021-0-01343, Artificial Intelligence Graduate School Program (Seoul National University)].

{\small
\bibliographystyle{ieee_fullname}
\bibliography{CVF-SID}
}

\appendixpageoff
\appendixtitleoff

\begin{appendices}

\title{Supplementary Material \textit{for} \\CVF-SID: Cyclic multi-Variate Function for \\ Self-Supervised Image Denoising by Disentangling  Noise from Image}

\author{Reyhaneh Neshatavar$^{1\ast}$ \qquad Mohsen Yavartanoo$^{1\ast}$ \qquad Sanghyun Son$^{1}$ \qquad Kyoung Mu Lee$^{1,2}$ 
\\$^{1}$Dept. of ECE \& ASRI, $^{2}$IPAI, Seoul National University, Seoul, Korea\\
{\tt\small \{reyhanehneshat,myavartanoo,thstkdgus35,kyoungmu\}@snu.ac.kr}}

\maketitle
\renewcommand{\thesection}{S1}
\section{Training on a Single Image}
\begin{figure}[t]
    \centering
    \captionsetup[subfigure]{labelformat=empty}
    \renewcommand{\wp}{\linewidth}
    
    \subfloat[]{\includegraphics[width=\wp, page=1]{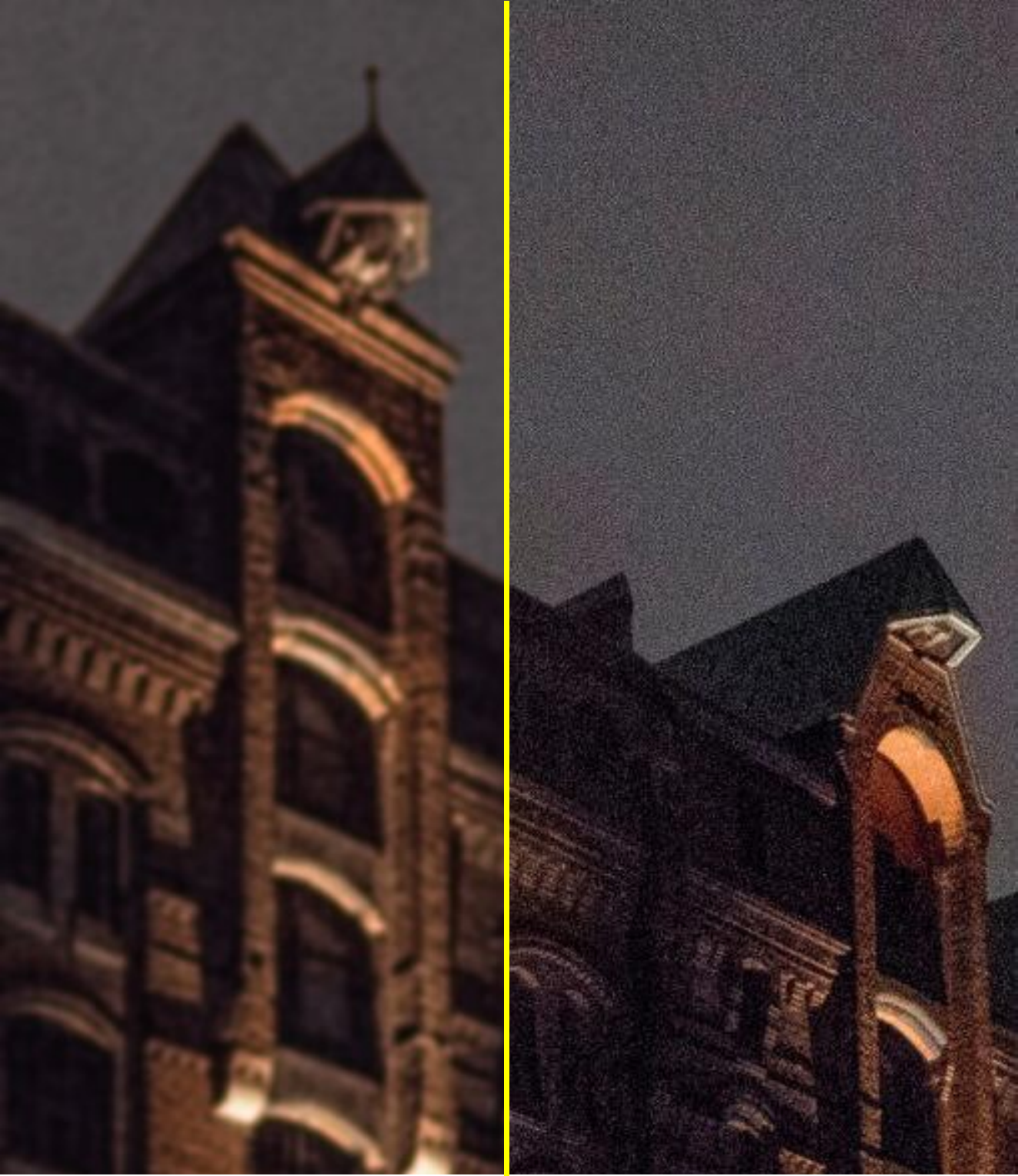}}
    \caption{
        \textbf{Restored clean image with only a single noisy image.}
    }
    \label{fig:single}
    \vspace{-4mm}
\end{figure}

We test our proposed CVF-SID on a practical case that uses only a single noisy image.
Specifically, we train our method in a self-supervised manner and apply it to a real-world input.
Figure~\ref{fig:single} demonstrates that our CVF-SID can learn to denoise without any other external examples.
%
During the training, we randomly crop patches to construct a mini-batch as we describe in Section~\textcolor{red}{4.2} in our main manuscript.
%
%
The denoising result shows that the proposed approach does not rely on a large-scale dataset but can be learned to denoise from a single image.

\renewcommand{\thesection}{S2}
\section{Network architecture details}
Our clean image generator consists of 16 sequentially $3\times3$ convolutional layers with the sizes of 64 and the padding size 1. 
Each convolution layer is followed by ReLU non-linear activation function.
Finally, a $1\times1$ convolutional layer generates the RGB clean image.

Our noise generator includes ten $3\times3$ convolutional layers with the sizes of 64 and padding size 1, followed by the ReLU activation function.
Then each branch of signal-dependent and signal-independent noise generators containing three $3\times3$ convolutional layers with ReLU activation function and one $1\times1$ convolutional layers is applied separately to generate the signal-dependent and signal-independent noise maps $\hat{N}_d$ and $\hat{N}_i$, respectively.

All convolutional weights and biases are initialized with Xavier uniform and constant 0, respectively.
To reduce the padding effects, we apply reflection padding of size 20 to each side of an input image and crop the output image to obtain the image with the original size. Moreover, to guarantee that the noise maps are zeros-mean, we subtract them with their channel-wise average.

\renewcommand{\thesection}{S3}
\section{Training Schemes}
\begin{figure*}
    \centering
    \captionsetup[subfigure]{labelformat=empty}
    \renewcommand{\wp}{0.195 \linewidth}
    \begin{minipage}[t]{0.99 \linewidth}
        \subfloat[\scriptsize \centering $I_n$]{\includegraphics[width=\wp, page=1]{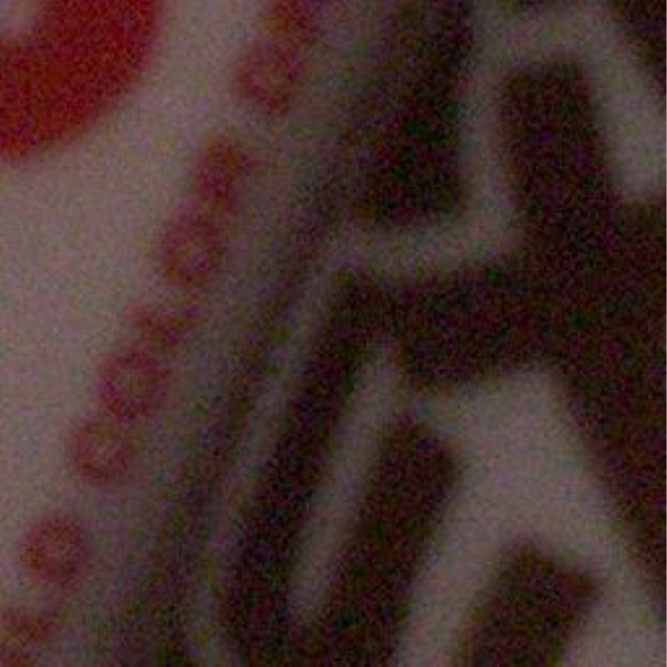}}
        \hfill
        \subfloat[\scriptsize \centering $\hat{I}_c (T)$ \protect\linebreak ($39.64$dB)]{\includegraphics[width=\wp, page=2]{figures/supp/supp_228.pdf}}
        \hfill
        \subfloat[\scriptsize \centering $\hat{I}_c (S)$ \protect\linebreak ($39.84$dB)]{\includegraphics[width=\wp, page=3]{figures/supp/supp_228.pdf}}
        \hfill
        \subfloat[\scriptsize \centering $\hat{I}_c (S^2)$ \protect\linebreak (\textbf{$\textbf{40.12}$dB})]{\includegraphics[width=\wp, page=4]{figures/supp/supp_228.pdf}}
        \hfill
        \subfloat[\scriptsize \centering $I_c$]{\includegraphics[width=\wp, page=5]{figures/supp/supp_228.pdf}}
        \\
        \\
        \subfloat[\scriptsize \centering $I_n$]{\includegraphics[width=\wp, page=1]{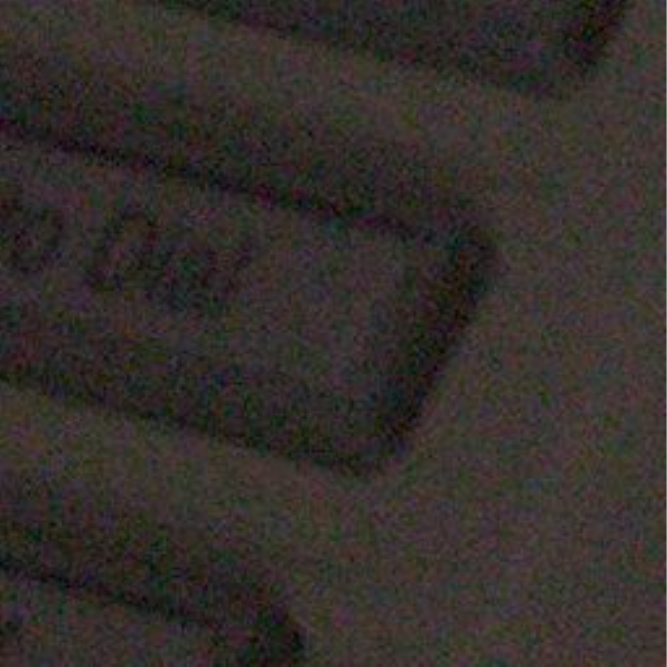}}
        \hfill
        \subfloat[\scriptsize \centering $\hat{I}_c (T)$ \protect\linebreak ($40.46$dB)]{\includegraphics[width=\wp, page=2]{figures/supp/supp_355.pdf}}
        \hfill
        \subfloat[\scriptsize \centering $\hat{I}_c (S)$ \protect\linebreak ($41.22$dB)]{\includegraphics[width=\wp, page=3]{figures/supp/supp_355.pdf}}
        \hfill
        \subfloat[\scriptsize \centering $\hat{I}_c (S^2)$ \protect\linebreak (\textbf{$\textbf{41.95}$dB})]{\includegraphics[width=\wp, page=4]{figures/supp/supp_355.pdf}}
        \hfill
        \subfloat[\scriptsize \centering $I_c$]{\includegraphics[width=\wp, page=5]{figures/supp/supp_355.pdf}}
        \\
        \subfloat[\scriptsize \centering $I_n$]{\includegraphics[width=\wp, page=1]{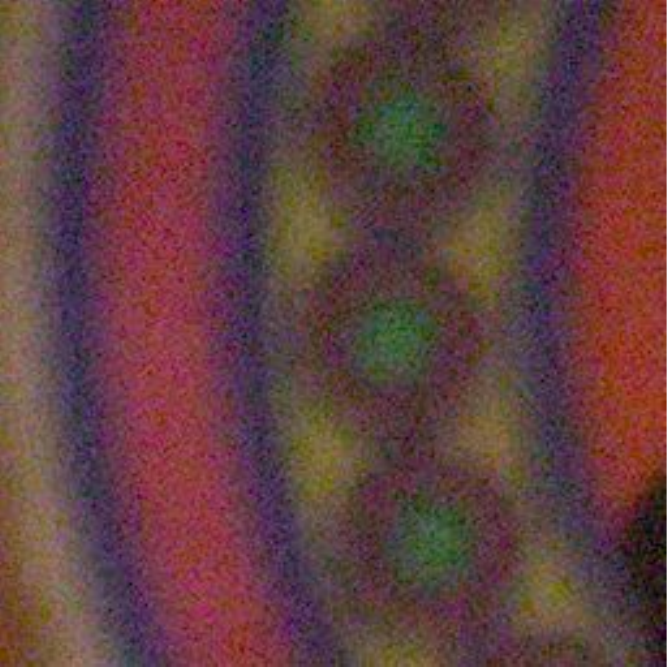}}
        \hfill
        \subfloat[\scriptsize \centering $\hat{I}_c (T)$ \protect\linebreak ($37.81$dB)]{\includegraphics[width=\wp, page=2]{figures/supp/supp_530.pdf}}
        \hfill
        \subfloat[\scriptsize \centering $\hat{I}_c (S)$ \protect\linebreak ($38.42$dB)]{\includegraphics[width=\wp, page=3]{figures/supp/supp_530.pdf}}
        \hfill
        \subfloat[\scriptsize \centering $\hat{I}_c (S^2)$ \protect\linebreak (\textbf{$\textbf{38.88}$dB})]{\includegraphics[width=\wp, page=4]{figures/supp/supp_530.pdf}}
        \hfill
        \subfloat[\scriptsize \centering $I_c$]{\includegraphics[width=\wp, page=5]{figures/supp/supp_530.pdf}}
        \\
        \subfloat[\scriptsize \centering $I_n$]{\includegraphics[width=\wp, page=1]{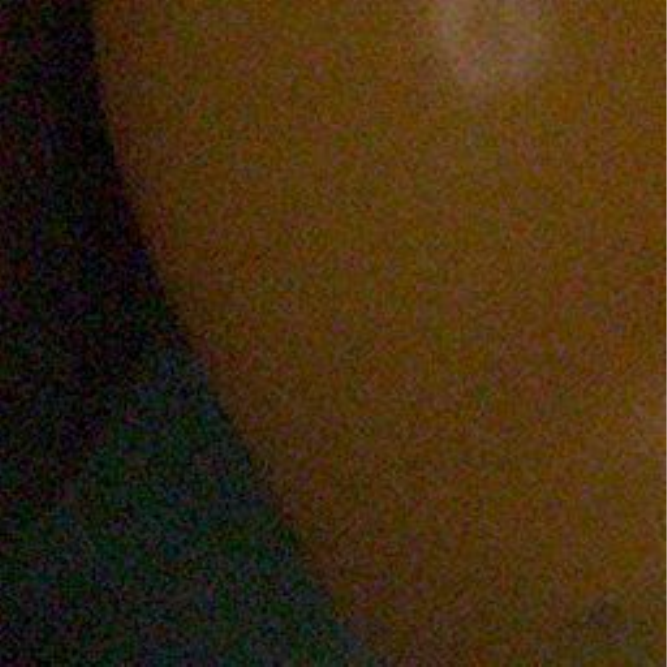}}
        \hfill
        \subfloat[\scriptsize \centering $\hat{I}_c (T)$ \protect\linebreak ($36.33$dB)]{\includegraphics[width=\wp, page=2]{figures/supp/supp_643.pdf}}
        \hfill
        \subfloat[\scriptsize \centering $\hat{I}_c (S)$ \protect\linebreak ($36.81$dB)]{\includegraphics[width=\wp, page=3]{figures/supp/supp_643.pdf}}
        \hfill
        \subfloat[\scriptsize \centering $\hat{I}_c (S^2)$ \protect\linebreak (\textbf{$\textbf{37.19}$dB})]{\includegraphics[width=\wp, page=4]{figures/supp/supp_643.pdf}}
        \hfill
        \subfloat[\scriptsize \centering $I_c$]{\includegraphics[width=\wp, page=5]{figures/supp/supp_643.pdf}}
        \\
        \subfloat[\scriptsize \centering $I_n$]{\includegraphics[width=\wp, page=1]{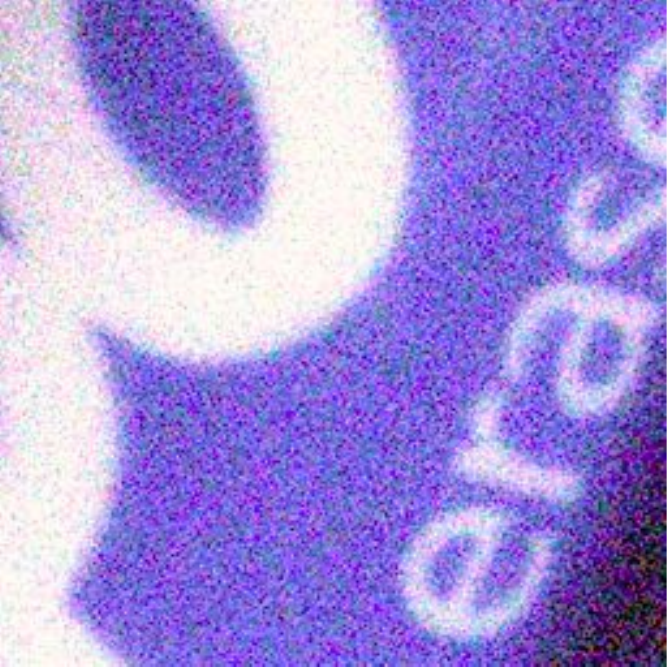}}
        \hfill
        \subfloat[\scriptsize \centering $\hat{I}_c (T)$ \protect\linebreak ($32.25$dB)]{\includegraphics[width=\wp, page=2]{figures/supp/supp_316.pdf}}
        \hfill
        \subfloat[\scriptsize \centering $\hat{I}_c (S)$ \protect\linebreak ($32.79$dB)]{\includegraphics[width=\wp, page=3]{figures/supp/supp_316.pdf}}
        \hfill
        \subfloat[\scriptsize \centering $\hat{I}_c (S^2)$ \protect\linebreak (\textbf{$\textbf{32.80}$dB})]{\includegraphics[width=\wp, page=4]{figures/supp/supp_316.pdf}}
        \hfill
        \subfloat[\scriptsize \centering $I_c$]{\includegraphics[width=\wp, page=5]{figures/supp/supp_316.pdf}}
    \end{minipage}
    \vspace{-2mm}
    \caption{
        \textbf{Visual comparisons of the predicted clean images from the SIDD validation dataset between different training schemes (T, S, and S$^2$).}
        We also provide PSNR \wrt ground-truth images.
    }
    \label{fig:supp_val}
\end{figure*}

Figure~\ref{fig:supp_val} provides visual comparisons between different training schemes \textbf{T}, \textbf{S}, and $\textbf{S}^2$ used to train our CVF-SID method.
%
%
\begin{figure*}
    \captionsetup[subfigure]{labelformat=empty}
    \renewcommand{\wp}{0.12 \linewidth}
    \subfloat{\includegraphics[width=\wp, page=1]{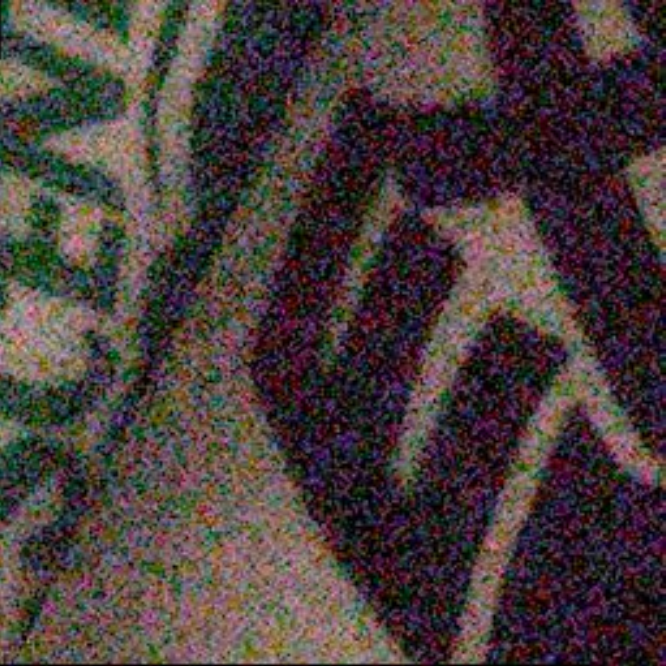}}
    \hfill
    \subfloat{\includegraphics[width=\wp, page=2]{figures/supp/supp_aug_182.pdf}}
    \hfill
    \subfloat{\includegraphics[width=\wp, page=3]{figures/supp/supp_aug_182.pdf}}
    \hfill
    \subfloat{\includegraphics[width=\wp, page=4]{figures/supp/supp_aug_182.pdf}}
    \hfill
    \subfloat{\includegraphics[width=\wp, page=5]{figures/supp/supp_aug_182.pdf}}
    \hfill
    \subfloat{\includegraphics[width=\wp, page=6]{figures/supp/supp_aug_182.pdf}}
    \hfill
    \subfloat{\includegraphics[width=\wp, page=7]{figures/supp/supp_aug_182.pdf}}
    \hfill
    \subfloat{\includegraphics[width=\wp, page=8]{figures/supp/supp_aug_182.pdf}}
    \\
    \subfloat{\includegraphics[width=\wp, page=1]{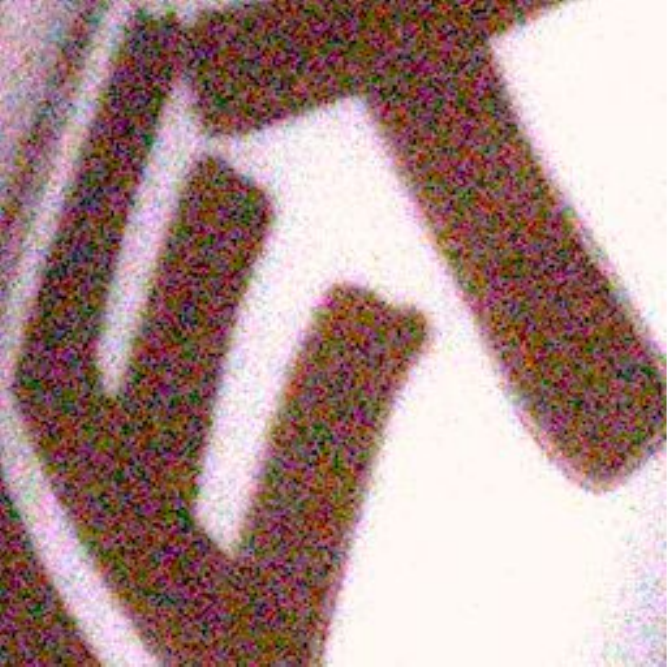}}
    \hfill
    \subfloat{\includegraphics[width=\wp, page=2]{figures/supp/supp_aug_292.pdf}}
    \hfill
    \subfloat{\includegraphics[width=\wp, page=3]{figures/supp/supp_aug_292.pdf}}
    \hfill
    \subfloat{\includegraphics[width=\wp, page=4]{figures/supp/supp_aug_292.pdf}}
    \hfill
    \subfloat{\includegraphics[width=\wp, page=5]{figures/supp/supp_aug_292.pdf}}
    \hfill
    \subfloat{\includegraphics[width=\wp, page=6]{figures/supp/supp_aug_292.pdf}}
    \hfill
    \subfloat{\includegraphics[width=\wp, page=7]{figures/supp/supp_aug_292.pdf}}
    \hfill
    \subfloat{\includegraphics[width=\wp, page=8]{figures/supp/supp_aug_292.pdf}}
    \\
    \subfloat[\scriptsize \centering $I_n$]{\includegraphics[width=\wp, page=1]{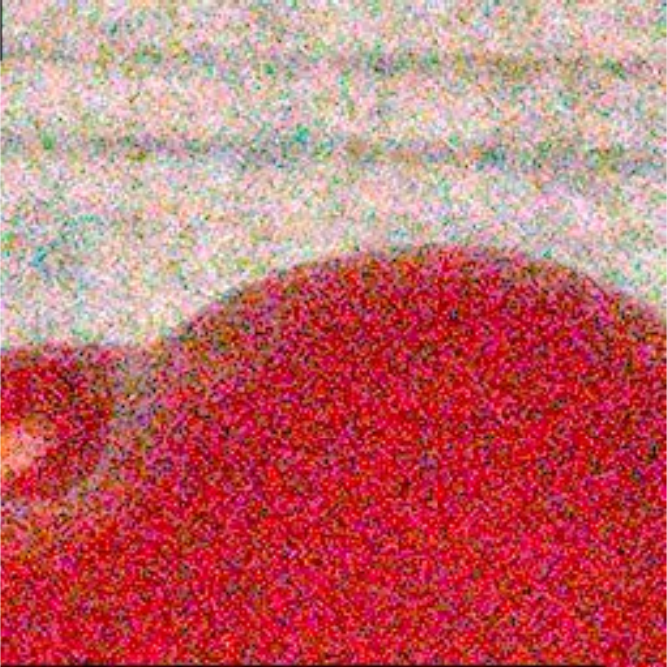}}
    \hfill
    \subfloat[\scriptsize \centering $\hat{I}_c+\hat{N}_i$]{\includegraphics[width=\wp, page=2]{figures/supp/supp_aug_736.pdf}}
    \hfill
    \subfloat[\scriptsize \centering $\hat{I}_c-\hat{N}_i$]{\includegraphics[width=\wp, page=3]{figures/supp/supp_aug_736.pdf}}
    \hfill
    \subfloat[\scriptsize \centering $\hat{I}_c-\hat{I}^{\gamma}_c\hat{N}_d$]{\includegraphics[width=\wp, page=4]{figures/supp/supp_aug_736.pdf}}
    \hfill
    \subfloat[\scriptsize \centering $\hat{I}_c+\hat{I}^{\gamma}_c\hat{N}_d+\hat{N}_i$]{\includegraphics[width=\wp, page=5]{figures/supp/supp_aug_736.pdf}}
    \hfill
    \subfloat[\scriptsize \centering $\hat{I}_c+\hat{I}^{\gamma}_c\hat{N}_d-\hat{N}_i$]{\includegraphics[width=\wp, page=6]{figures/supp/supp_aug_736.pdf}}
    \hfill
    \subfloat[\scriptsize \centering $\hat{I}_c-\hat{I}^{\gamma}_c\hat{N}_d+\hat{N}_i$]{\includegraphics[width=\wp, page=7]{figures/supp/supp_aug_736.pdf}}
    \hfill
    \subfloat[\scriptsize \centering $\hat{I}_c-\hat{I}^{\gamma}_c\hat{N}_d-\hat{N}_i$]{\includegraphics[width=\wp, page=8]{figures/supp/supp_aug_736.pdf}}
    \caption{
        \textbf{Generated synthetic noisy images by our proposed self-supervised augmentation strategy on the SIDD validation dataset.}
    }
    \label{fig:supp_aug}
\end{figure*}
\begin{figure*}[t]
    \captionsetup[subfigure]{labelformat=empty}
    \renewcommand{\wp}{0.48\linewidth}
    
    \subfloat[]{\includegraphics[width=\wp, page=1]{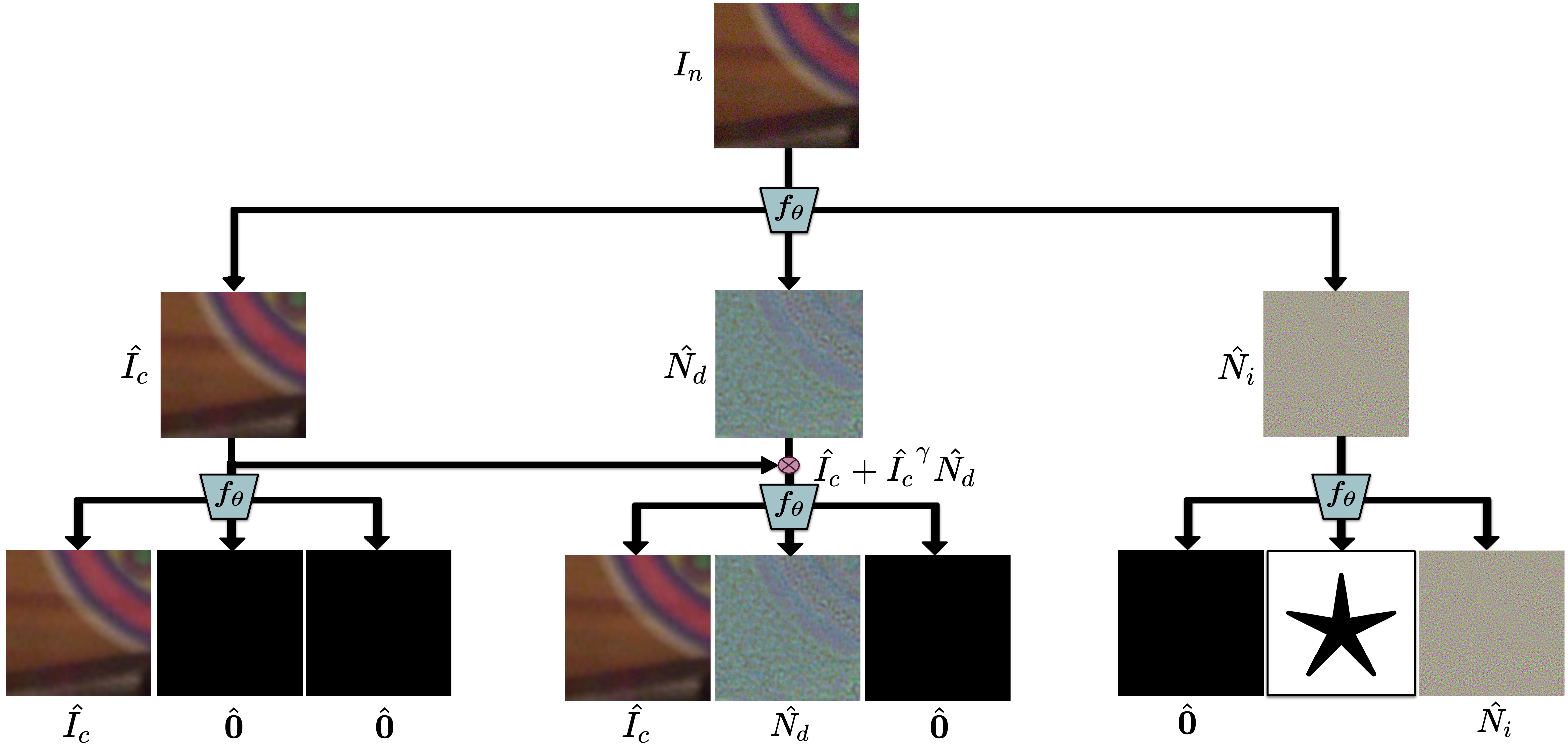}}
    \hfill
    \subfloat[]{\includegraphics[width=\wp, page=2]{figures/supp/supp_framework.pdf}}
    \\
    \subfloat[]{\includegraphics[width=\wp, page=3]{figures/supp/supp_framework.pdf}}
    \hfill
    \subfloat[]{\includegraphics[width=\wp, page=4]{figures/supp/supp_framework.pdf}}
    \caption{
        \textbf{Decomposition results of our CVF-SID method on the SIDD validation dataset with training scheme (S).}
    }
    \label{fig:supp_frame}
\end{figure*}
Results demonstrate that our method achieves slightly better results with training scheme $\textbf{S}$ than \textbf{T} for all images because the discrepancy between training and test images can be minimized as discussed in Section~\textcolor{red}{4.3} in our main manuscript.
%
Furthermore, we show that in some cases, the training scheme \textbf{S} is sufficient to remove the noise, and we do not need to double-train model (\textbf{S$^2$}) to further achieve better performance, as shown in fifth row of Figure~\ref{fig:supp_val}.
%

\renewcommand{\thesection}{S4}
\section{Cascading additional models}
We further apply three (\textbf{S$^3$}) and four (\textbf{S$^4$}) CVF-SID, successively, to the original noisy images from the SIDD validation dataset.  
The results show the improvement of model performance to some extent but saturate at $\mathbf{S}^3$ as follows:
%
\renewcommand{\thetable}{S1}
\begin{table}[h]
    \vspace{-2mm}
    \centering
    \small
    \setlength\tabcolsep{1.5pt} 
    \begin{tabularx}{\linewidth}{>{\raggedright\arraybackslash}X c c c c}
        \toprule
        \textbf{Method} & \textbf{S} & $\mathbf{S^2}$ & $\mathbf{S^3}$ & $\mathbf{S^4}$ \\
        \midrule
           \textbf{PSNR/SSIM} & 34.67/0.943 & 34.81/\textbf{0.944} & {\bf 34.84}/0.943 & {\bf 34.84}/0.942 \\
        \bottomrule
    \end{tabularx}
    \vspace{-2mm}
    \caption{
        \textbf{Quantitative comparison on the number of cascading models on sRGB images in the SIDD validation dataset.}
    }
    \vspace{-4mm}
    \label{tab:cascade}
\end{table} 

\renewcommand{\thesection}{S5}
\section{Augmentations}
In Figure~\ref{fig:supp_aug}, we visualize various synthesized images from  our self-supervised augmentation strategy described in Section~\textcolor{red}{3.2} in our main manuscript.
%
%
%
We note that our augmentation strategy can generate several real-world noisy images from only a single noisy image without requiring additional information.

We also analyze the effect of the hyperparameter $\lambda_\text{aug}$ on the SIDD validation dataset as follows:

\renewcommand{\thetable}{S2}
\begin{table}[h]
    \vspace{-2mm}
    \centering
    \small
    \setlength\tabcolsep{1.5pt} 
    \begin{tabularx}{\linewidth}{>{\raggedright\arraybackslash}X c c c c}
        \toprule
        \textbf{$\lambda_\text{aug}$} & 0 & 0.01 & 0.1 & 1 \\
        \midrule
           \textbf{PSNR/SSIM} & 34.43/0.942 & 34.53/0.942 & \textbf{34.67}/\textbf{0.943} & 34.55/0.936 \\
        \bottomrule
    \end{tabularx}
    \vspace{-2mm}
    \caption{
        \textbf{Quantitative comparison on hyperparameter ${\bf\lambda_{aug}}$ on sRGB images in the SIDD validation dataset.}
    }
    \vspace{-4mm}
    \label{tab:gamma}
\end{table} 
\renewcommand{\thesection}{S6}
\section{Decomposition results}
Figure~\ref{fig:supp_frame} shows how our CVF-SID can effectively disentangle the clean image, signal-dependent, and signal-independent noises from a noisy input image.
%
%
The results demonstrate that CVF-SID is successfully learned to satisfy our constraints on its outputs, which are described in Section~\textcolor{red}{3.2}.
%
For example, when we feed the initially predicted noise-free image $\hat{I}_c$ again to the network $f_\theta$, we can get the same clean image for $f_\theta^\text{clean} ( \hat{I}_c )$ and zeros for the corresponding noise maps.
%
One limitation of our method is that the predicted clean image $\hat{I}_c$ and the signal-dependent noise map $\hat{N}_d$ are not completely independent, which is opposed to our assumption mentioned in Section~\textcolor{red}{3.2}.  
This contradiction can be due to considering a fixed correlation parameter $\gamma$, which may vary per image for a real-world scenario.



\end{appendices}

\end{document}